\pdfoutput=1
\documentclass[11pt]{article}
\usepackage[preprint]{acl}
\usepackage{times}
\usepackage{latexsym}
\usepackage[T1]{fontenc}
\usepackage[utf8]{inputenc}
\usepackage{microtype}
\usepackage{inconsolata}
\usepackage{graphicx}
\title{Brittle Minds, Fixable Activations:\\Understanding Belief Representations in Language Models}
\author{
Matteo Bortoletto \quad 
Constantin Ruhdorfer \quad 
Lei Shi \quad
Andreas Bulling \\
University of Stuttgart, Germany \\
\texttt{matteo.bortoletto@vis.uni-stuttgart.de} 
}
\usepackage{microtype}
\usepackage[utf8]{inputenc} 
\usepackage[T1]{fontenc}    
\usepackage{hyperref}       
\usepackage{url}            
\usepackage{booktabs}       
\usepackage{amsfonts}       
\usepackage{nicefrac}       
\usepackage{microtype}      
\usepackage{enumitem}
\setlist[itemize]{leftmargin=*}
\usepackage{graphicx}
\usepackage{subcaption}
\usepackage{bm}
\usepackage{amsmath}
\usepackage{physics}
\usepackage{wrapfig}
\usepackage{cleveref}
\usepackage{tcolorbox}
\usepackage{kotex}
\usepackage{multirow}
\usepackage{colortbl} 
\usepackage{calc}
\definecolor{Bittersweet}{rgb}{1.0, 0.44, 0.37}
\definecolor{CadetBlue}{rgb}{0.37, 0.62, 0.63}
\definecolor{Plum}{rgb}{0.56, 0.27, 0.52}
\definecolor{PineGreen}{rgb}{0.0, 0.47, 0.44}
\definecolor{cmagenta}{HTML}{99004D}
\definecolor{cblue}{HTML}{007FFF}
\definecolor{cred}{HTML}{FF0000}
\definecolor{ablue}{HTML}{61AAF2}
\definecolor{ared}{HTML}{CC5247}

\newtcolorbox[auto counter]{example}[3]{
    label=#1,
    title=Example~\thetcbcounter:~#3,
    colback=#2!5!white,
    colframe=#2!75!black,
}
\renewenvironment{quote}{%
  \list{}{%
    \leftmargin0.5cm   
    \rightmargin\leftmargin
    \topsep0.1cm
  }
  \item\relax
}
{\endlist}

\begin{document}
\maketitle
\begin{abstract}
    Despite growing interest in Theory of Mind (ToM) tasks for evaluating language models (LMs), little is known about how LMs \textit{internally represent mental states} of self and others. 
Understanding these internal mechanisms is critical -- not only to move beyond surface-level performance, but also for model alignment and safety, where subtle misattributions of mental states may go undetected in generated outputs.
In this work, we present the first systematic investigation of belief representations in LMs by probing models across different scales, training regimens, and prompts -- using control tasks to rule out confounds. 
Our experiments provide evidence that both model size and fine‑tuning substantially improve LMs' internal representations of others' beliefs, which are structured -- not mere by-products of spurious correlations -- yet brittle to prompt variations. 
Crucially, we show that these representations can be strengthened: targeted edits to model activations can correct wrong ToM inferences.
\end{abstract}
\section{Introduction}

Language models (LMs) trained on next token prediction have demonstrated impressive capabilities across various tasks, spanning coding, math, and embodied interaction
~\citep{wei2022emergent, bubeck2023sparks}. 
As these models are designed with the ultimate goal of collaborating with humans, it becomes imperative that they complement these skills with an understanding of humans. 
Core to this understanding is \textit{Theory of Mind} (ToM) -- the ability to attribute mental states to oneself and others~\citep{premack1978does}. 
ToM is essential for effective communication and cooperation with other agents, facilitating interaction and learning from feedback and demonstrations~\citep{saha2023can}.
Given its significance, computational ToM has emerged as 
a key capability when evaluating cutting-edge LMs~\citep{ma2023towards, shapira2024clever, chen2025theory}.

\begin{figure}[t]
    \centering
    \includegraphics[width=\linewidth]{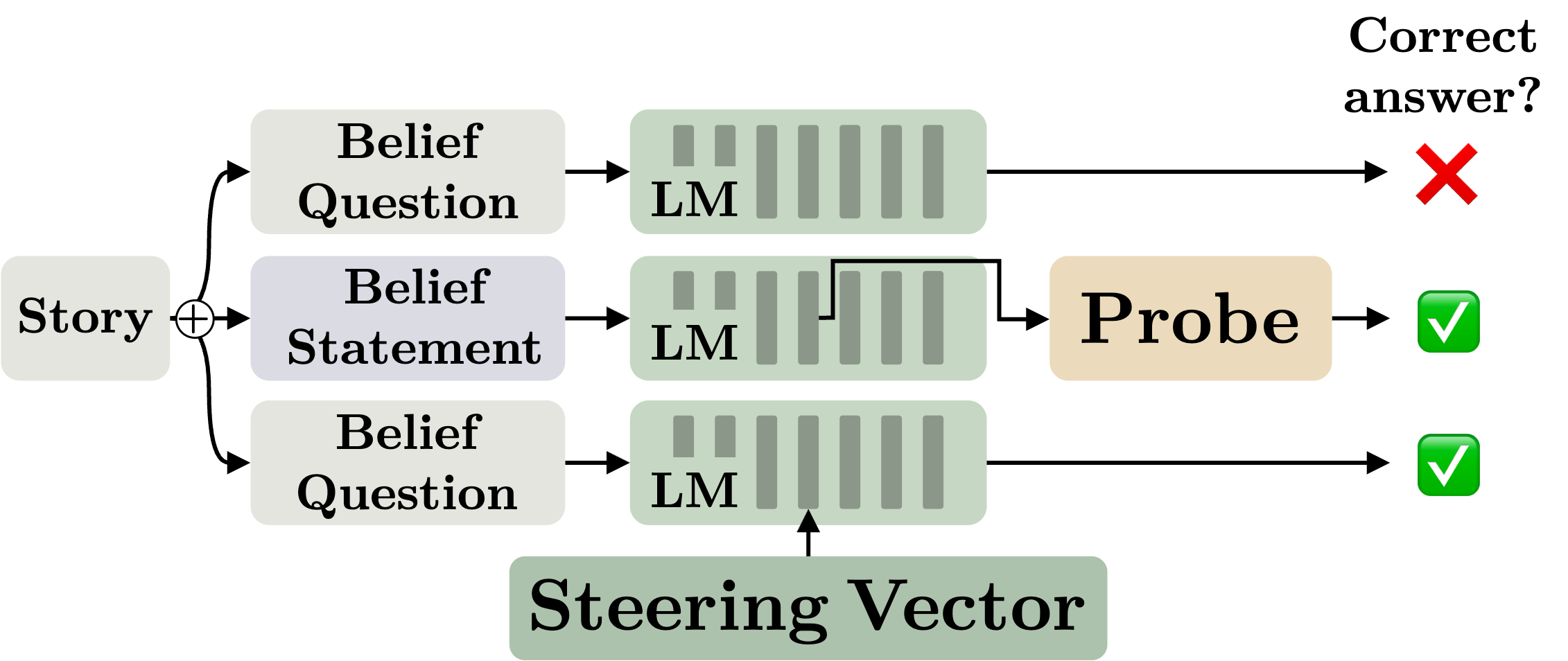}
    \caption{ToM tasks are challenging for LMs, but correct predictions can sometimes be recovered by \textit{probing} their internal representations. 
    We study how internal representations of beliefs of self and others emerge in 12 LMs, and show that these representations are structured yet brittle to prompts, and can be strengthened with a steering vector to fix incorrect ToM inferences.
    \vspace{-0.6cm}
    }
    \label{fig:teaser}
\end{figure}

Despite the improved performance on ToM benchmarks compared to earlier models, modern LMs are still far from perfect \citep{sap2022neural}.
Text generated by LMs often contains errors that limit their performance on ToM tasks. 
\citet{zhu2024language} showed that \textit{probing} LMs’ internal representations can sometimes recover correct belief inferences, with models like Mistral-7B-Instruct \cite{jiang2023mistral} and DeepSeek-7B-Chat \cite{bi2024deepseek} capturing beliefs from both their own and others' perspectives. 
While promising, this remains a preliminary step: 
it examines only single-sized, fine-tuned models, leaves possible confounds uncontrolled, and ignores how subtle changes in prompting affect belief representations.
As a result, we still lack a clear understanding of how internal belief representations differ across models, whether they reflect true ToM or spurious patterns, and how robust they are to prompts.

To address these gaps, we pose four key research questions and present evidence for each. 
We begin by studying how emergence scales across models:
\begin{description}[leftmargin=!,itemsep=0pt,topsep=2pt]
    \item[RQ1] \textit{Do internal belief representations \textbf{emerge} similarly in different LMs, and are they \textbf{affected} by model size and training regime?} 
\end{description}
Finding training regimes or scales that are more conducive to belief reasoning can guide future model development toward more reliable ToM behaviour.
However, it is also crucial to verify if representations are structured, indicating genuine modelling of mental states, or spurious:
\begin{description}[leftmargin=!,itemsep=0pt,topsep=2pt]
    \item[RQ2] \textit{Are LMs' internal belief representations \textbf{structured} or the result of spurious correlations?}
\end{description}
This distinction is essential for determining if representations reflect a genuine understanding of beliefs or only exploit statistical patterns that happen to correlate with correct answers in the training data. 
This is also crucial for alignment and safety, as misaligned mental state attributions may not appear overtly in text -- leading to false signals of understanding. 
Equally important is that models can maintain robust belief attributions:
\begin{description}[leftmargin=!,itemsep=0pt,topsep=2pt]
    \item[RQ3] \textit{Are LMs' internal belief representations \textbf{robust}?}
\end{description}
Fragile representations may break under slight variations, leading to inconsistent or unsafe behaviour in real-world applications involving social reasoning or user interaction.
Strengthening these representations, then, offers a promising path toward improving their reliability:
\begin{description}[leftmargin=!,itemsep=0pt,topsep=2pt]
    \item[RQ4] \textit{Can we \textbf{strengthen} LMs' internal belief representations to improve their performance?}
\end{description}

To answer these research questions, we perform probing and activation editing experiments using \textbf{12 LMs} (\Cref{fig:teaser}).
We first compare base models with those fine-tuned via SFT and/or RLHF \cite{ouyang2022training}(\textbf{RQ1}), finding that belief representations emerge in consistent patterns across models, improve with model size, and -- especially in smaller models -- benefit significantly from fine-tuning.
To provide evidence that LMs' belief representations are structured (\textbf{RQ2}), we show that (1) probes trained on randomly permuted labels perform at chance -- confirming selectivity, and (2) probes trained on top-$k$ principal components still recover most accuracy for $k \ll d_{model}$.
Next, we test robustness (\textbf{RQ3}) using varied prompts. 
Surprisingly, semantically neutral changes can reduce accuracy, revealing that representations of others' beliefs are brittle to prompts.
However, we show that it is possible to strengthen models' representation by using contrastive activation addition \cite[CAA]{rimsky2023steering}, obtaining significant performance improvements across different ToM tasks (\textbf{RQ4}). 

In summary, our work makes the following contributions:
\begin{enumerate}[leftmargin=0.5cm,itemsep=0cm,topsep=0pt]
    \item We provide extensive probing experiments across 12 LMs, suggesting that representations of others' beliefs improve with size and fine-tuning, and that these representations are structured yet brittle to prompt variations.
    \item We show that we can strengthen models' representations by using contrastive activation addition and improve their ToM performance. 
\end{enumerate}
\section{Related Work}

\paragraph{Machine Theory of Mind}
Theory of mind has been studied in AI for more than a decade \citep{baker2009action, rabinowitz2018machine, bara2021mindcraft, bortoletto2024limits, bortoletto2024explicit, bortoletto2024neural}.
Various benchmarks have been proposed, aiming to measure LMs' ability to understand and reason about the beliefs, goals, and intentions of others \citep{le2019revisiting, he2023hi, kim2023fantom, gandhi2024understanding, xu2024opentom, tan2024phantom, sclar2023minding, ma2023towards, wu2023coke}.
Additionally, efforts have been made to enhance LMs' ToM through prompting techniques \citep{zhou2023far, moghaddam2023boosting, wilf2023think}. 
Our work dives deeper into LMs' internal belief representations, offering a broader insight into these mechanisms that go beyond surface-level performance.

\paragraph{Probing Neural Representations}

Initially proposed by \citet{alain2016understanding}, probing is a widely used method for determining if models represent particular features or concepts. 
In the realm of LMs, numerous works used probing to demonstrate that these models acquire rich linguistic representations -- spanning 
semantic concepts such as syntactic categories, dependency relations, co-reference, and word meaning \citep{conneau2018you,tenney2018you,tenney2019bert,rogers2021primer,li2021implicit,hernandez2021low,marks2023geometry,liu2023cognitive}.
A separate line of work explored if LMs possess a \textit{world model} \citep{li2021implicit, abdou2021language, patel2022mapping, li2023emergent, nanda2023emergent}.
An emergent line of work that is relevant to our work used probing to explore if LMs have \textit{agent models}, for example, if they can represent beliefs of self and others \citep{zhu2024language, bortoletto2024limits}. 
In this work, we contribute with extensive experiments that  characterise models’ representations of beliefs along different axes: emergence, structure, robustness, and steerability.

\paragraph{Prompt Analysis}
Previous work has shown that LMs are vulnerable to prompt alterations like token deletion or reordering \citep{ishibashi2023evaluating}, biased or toxic prompts \citep{shaikh2023second} and similarity to training data \citep{razeghi2022impact}. 
Other works have shown the importance of input-output format \citep{min2022rethinking} and of demonstration example ordering for few-shot performance \citep{zhao2021calibrate, lu2022fantastically, zhou2023leasttomost}.
In this work, \textit{we shift our focus from analysing how sensitive model outputs are to how model representations change} \cite{gurnee2023language}. 
In particular, we explore for the first time the effect of prompt variations on how models internally represent mental states.

\paragraph{Activation Editing}
Activation editing has emerged as a way to influence model behaviour without any additional fine-tuning \citep{li2023emergent, hernandez2023inspecting}. 
One notable method in this domain is inference-time intervention \cite[ITI]{li2023inferencetime}, which
involves training linear probes on contrastive question-answering datasets to identify ``truthful'' attention heads and then shifting their activations during inference along the identified truthful directions. 
In contrast, activation addition \cite[AA]{turner2023activation} and contrastive activation addition \cite[CAA]{rimsky2023steering} generate \textit{steering vectors} by only using LMs' activations.
\citeauthor{zhu2024language} has used ITI to show that it is possible to manipulate LMs' internal representations of mental states. 
In this work, we show that using CAA can further improve LMs' ToM capabilities while eliminating the need for a fine-grained search over attention heads.
\section{Experimental Setup}

\subsection{Probing}
We linearly decode belief status from the perspective of different agents by using probing \citep{alain2016understanding}. 
Probing involves localising specific concepts in a neural model by training a simple classifier (called a \textit{probe}) on model activations to predict a target label associated with the input data.
To provide a formal definition, we adopt a similar notation to the one introduced in \citep{belinkov2022probing}. 
Consider an \textit{original model} $f: x \mapsto \hat{y}$ that is trained on a dataset $\mathcal{D}^O = \{x^{(i)}, y^{(i)}\}$ to map input $x$ to output $\hat{y}$. 
Model performance is evaluated by some measure, denoted $\textsc{Perf}(f, \mathcal{D}^O)$.
A \textit{probe} $g_l: f_l(x) \mapsto \hat{z}$ maps intermediate representations of $x$ in $f$ at layer $l$ to some property $\hat{z}$, which is the label of interest.
The probe $g_l$ is trained on a \textit{probing dataset} $\mathcal{D}^P = \{x^{(i)}, z^{(i)}\}$ and evaluated using some performance measure $\textsc{Perf}(g_l, f, \mathcal{D}^O, \mathcal{D}^P)$.  
In our case, $f$ is an autoregressive language model that, given a sequence of tokens $x$, outputs a probability distribution over the token vocabulary to predict the next token in the sequence. 
Our probe is a logistic regression model $g_l: \hat{z} = Wa_l + b$ trained on neural activations $f_l(x) = a_l$ to predict binary belief labels $y=\{0,1\}$.

\subsection{Dataset} 
\begin{figure}[t]
    \centering
    \includegraphics[width=\linewidth]{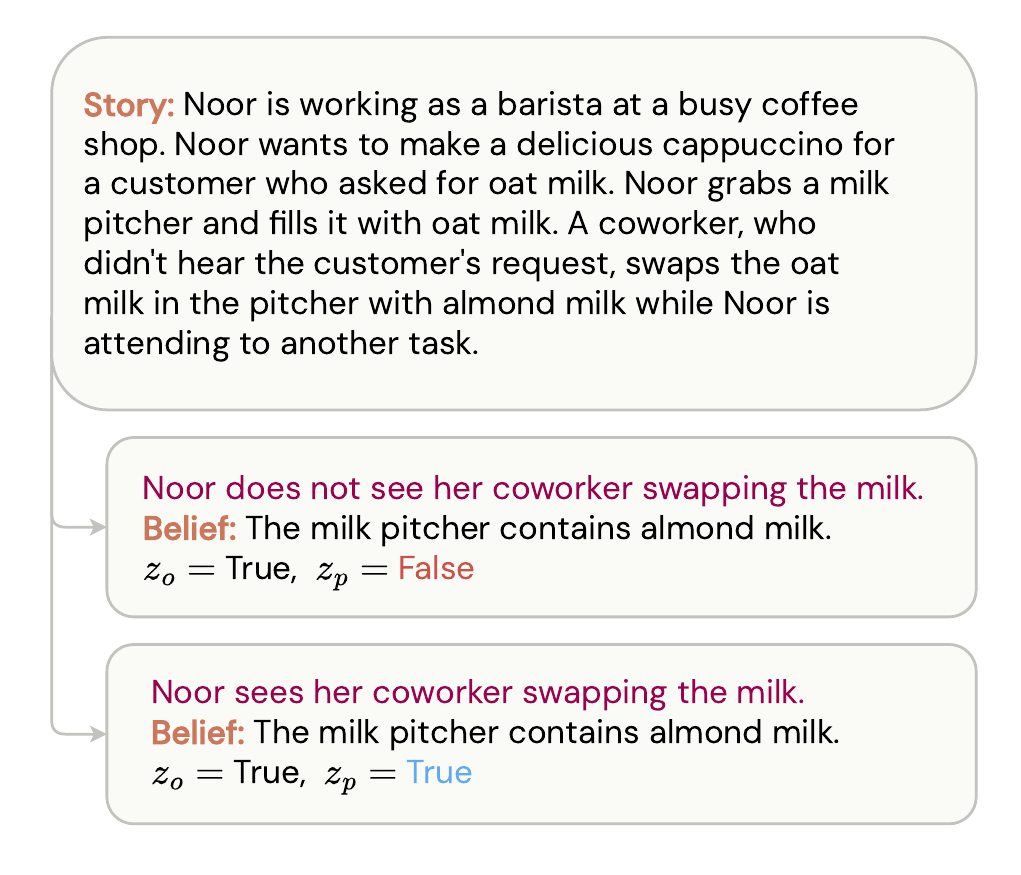}
    \caption{Example of false belief from our probing datasets. The labels $z_p$ and $z_o$ correspond to $\mathcal{D}_p^P$ and $\mathcal{D}_o^P$, respectively. By manipulating the protagonist's \textcolor{cmagenta}{percepts} after the causal event, we obtain two scenarios: \textcolor{ablue}{true belief} and \textcolor{ared}{false belief}.}
    \label{fig:probing-dataset}
\end{figure}

We use BigToM \citep{gandhi2024understanding}, a question-answering dataset constructed by populating causal templates and combining elements from these templates. 
Each causal template is set up with a \textit{context} and a description of the \textit{protagonist} (e.g. \textit{``Noor is working as a barista [\dots]''}, see Story in \Cref{fig:probing-dataset}), a \textit{desire} (\textit{``Noor wants to make a cappuccino''}), a \textit{percept} (\textit{``Noor grabs a milk pitcher and fills it with oat milk''}), and a \textit{belief} (\textit{``Noor believes that the pitcher contains oat milk''}). 
The state of the world is changed by a \textit{causal event} (\textit{``A coworker swaps the oat milk in the pitcher with almond milk''}).
The dataset constructs different conditions by changing the percepts of the protagonist after the causal event, which will result in different beliefs. 
Similar to \citep{zhu2024language}, we focus on the \textit{Forward Belief} setting in which models have to infer the belief of the protagonist given the percepts of the causal event, $P(\mathrm{belief}|\mathrm{percepts})$. 
We report additional details in \Cref{app:bigtom}

\paragraph{Probing Datasets} 
We consider two probing datasets: $\mathcal{D}_p^P = \{x^{(i)}_p, z^{(i)}_p\}$, where the labels $z^{(i)}_p$ correspond to ground-truth beliefs from the \textit{protagonist} perspective, and $\mathcal{D}_o^P = \{x^{(i)}_o, z^{(i)}_o\}$, where the labels $z^{(i)}_o$ reflect the perspective of an omniscient \textit{oracle}.
$\mathcal{D}_p^P$ and $\mathcal{D}_o^P$ are built by pairing each story in BigToM with a belief statement, as shown in \Cref{fig:probing-dataset}.
After prompting the model with a story-belief pair $x$ we cache the residual stream activations $f_l(x)$ at the final token position for all residual streams (see \Cref{fig:probe}).

\subsection{Models} 
We study two families of LMs that offer us options in model sizes and fine-tuning: Pythia \citep{biderman2023pythia} and Llama-2 \citep{touvron2023llama} -- for a total of \textbf{12 models}. 
While Llama-2 offers ``chat'' versions first trained with SFT and then RLHF, Pythia's open-source training set \citep{gao2020pile} ensures that there is no data leakage\footnote{Llama-2 was released later than BigToM.}.
Additionally, we consider a SFT version of Pythia-6.9B trained on open-source instruction datasets \citep{wang2024far}, which we refer to as Pythia-6.9B-chat.\footnote{\url{https://huggingface.co/allenai/open-instruct-pythia-6.9b-tulu}}
We provide model details in \Cref{tab:models_details}.

\subsection{Probing Experiments}
\label{subs:probing-exp}
To study how LMs represent beliefs of self and others, we propose a set of extensive probing experiments across LMs that differ in architecture, size, and fine-tuning regime. 
We train probes on the residual stream, as it integrates information from both the attention and feed-forward components, potentially encoding richer representations.
Additionally, since the residual activations directly contribute to the final output predictions, probing them may better align with understanding the model's behaviour for downstream tasks.

\paragraph{Control Tasks}
Depending on the model dimension, the probes we train have a significant number of learnable parameters -- up to $16,385$ for Llama-2-70B.
This raises the concern that probes might learn to rely on irrelevant patterns in the data instead of capturing meaningful relationships. 
To account for the potential confounding effect of hidden state size, we include two controls.
First, following \citet{hewitt2019designing}, we train and evaluate probes on a version of $\mathcal{D}^P_p$ with randomly permuted labels -- thus removing real input-label relationships.
If a probe still performs well on the permuted data, this suggests it may be exploiting superficial correlations rather than capturing genuine structure.
Second, we effectively reduce the number of learnable parameters in the probes by projecting $\mathcal{D}_p^P$ and $\mathcal{D}_o^P$ onto their $k$ largest principal components using PCA before training.
This minimises the risk of the probes relying on spurious patterns in the data. 

\paragraph{Robustness Tests} 
Previous work left the impact of prompting on belief probing accuracy unexplored. 
Our second set of experiments aims to study whether belief representations are robust to different prompts.
Research on prompt robustness in language models focused mainly on revealing vulnerability to prompt alterations on \textit{downstream performance} \citep{min2022rethinking, ishibashi2023evaluating,shaikh2023second, leidinger2023language, sclar2024quantifying}. 
In contrast, we study how different prompt alterations influence \textit{probing performance}, i.e. models' internal representations. 
Unlike model outputs that are shaped by decoding strategies, which act as confounders, models' activations are more abstract and offer a better lens into how robust or brittle internal representations are.
We define four prompt variations:
\begin{itemize}[itemsep=0cm,topsep=2pt]
    \item \textit{Random}: Following \citet{gurnee2023language}, we add 10 random tokens to the belief statement.
    \item \textit{Misleading}: Each story is followed by two belief statements, one pertinent to the story and one randomly chosen from another.
    \item \textit{Time Specification}: The prompt specifies that the belief statement refers to the end of the story. 
    We include this variation because some belief statements can be true (false) at the story's beginning but false (true) at the end. 
    For example, consider the story in \Cref{fig:probing-dataset}: if Noor does not witness the swap, in the end, she will believe the pitcher contains almond milk ($z_p=\mathrm{True}$). 
    However, if the same belief is referred to the beginning of the story, then it is false ($z_p=\mathrm{False}$). 
    \item \textit{Initial Belief}: We explicitly reveal the protagonist's initial belief (e.g.\ \textit{``Noor believes that the pitcher contains oat milk''}) in the story to test whether it biases the representations of LMs.
\end{itemize}
While all maintain conceptual and semantic parity with the \textit{Original} prompt used in \citep{zhu2024language}, \textit{Random} and \textit{Misleading} are expected to negatively impact LMs' representations, while \textit{Time Specification} and \textit{Initial Belief} are supposed to have a positive influence. 
Robust representations of beliefs should exhibit minimal sensitivity to these alterations.
Our experiments compare probe accuracy across different model sizes, fine-tuning, and prompt variations. 
Examples of prompts are reported in \Cref{app:prompts}.

\begin{figure*}[t]
  \centering
  \includegraphics[width=\textwidth]{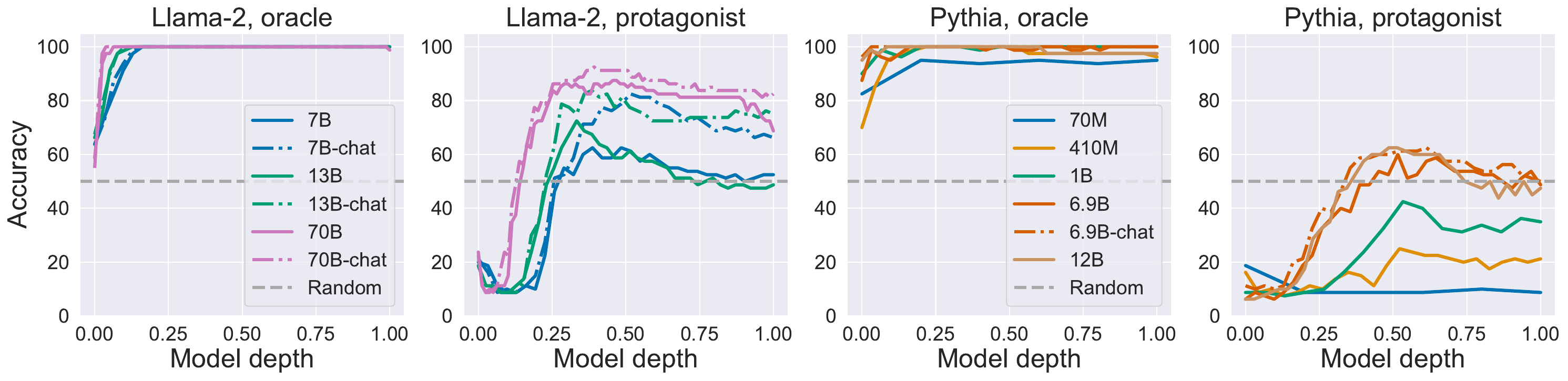}
  \caption{Belief probing accuracy
  show similar patterns across all models: \textit{oracle} belief representations generally form already in the first layers, while \textit{protagonist} belief representations emerge at the intermediate layers. 
  Moreover, probing accuracy increases with model size and, more crucially for smaller models, with fine-tuning.}
  \label{fig:size-ft}
\end{figure*}

\subsection{Activation Editing}
Prior work found that it is possible to manipulate models' representations of beliefs by using \cite[ITI]{li2023inferencetime}, and that such interventions can improve LMs' performance on ToM tasks. 
We take this further by asking whether a general ``belief vector'' can be distilled and \textit{injected} into the models' activations to  \textit{strengthen} their ToM abilities. 
To this end, we use contrastive activation addition \cite[CAA]{rimsky2023steering}, an extension of activation addition \cite[AA]{turner2023activation} that computes \textit{steering vectors} to control LMs' behaviour. 
Steering vectors are computed as the average difference in residual stream activations between pairs of positive and negative instances of a specific behaviour.
Formally, given a dataset $\mathcal{D}$ of triplets ($p$, $c_p$, $c_n$), where $p$ is a prompt, $c_p$ is a positive completion, and $c_n$ is a negative completion, CAA computes a \textit{mean difference} vector $v^{md}_l$ for layer $l$ as:
\begin{equation}
    v^{md}_l = \frac{1}{\abs{\mathcal{D}}} \sum_{p,c_p,c_n \in \mathcal{D}} a_l(p,c_p) - a_l(p,c_n)
\end{equation}
For example, in Figure \ref{fig:probing-dataset}, $p$ is the \textit{Story}, $c_p$ could be the true belief, and $c_n$ the false belief. 
During inference, these steering vectors are multiplied by an appropriate coefficient $\alpha$ and added at every token position of the generated text after the prompt.
CAA has two main advantages over ITI:
First, it eliminates the need to train probes, making it \textit{computationally cheap}. 
For example, for Llama2 70B, ITI needs to train 5,120 probes while CAA only needs to compute 80 vectors. 
Second, it operates at the residual stream level, making it easier to use than methods that intervene on specific attention heads like ITI. 
While CAA has been used to control alignment-relevant behaviour, such as hallucinations, refusal, and sycophancy \citep{rimsky2023steering}, we are the first to apply it to enhance LMs' ToM reasoning. 
The ``belief vectors'' (i.e. steering vectors) we obtain can be understood as isolating the direction in the LMs' latent space corresponding to taking the perspective of another agent. 
To evaluate both base and fine-tuned LMs, we rank their answers to the ToM questions according to $p_{LM}(a|q)$ \citep{petroni2019language}. 
For a fair comparison, we adopt the train/test \textit{Forward Belief} split used in \citep{zhu2024language} to compute and evaluate the steering vectors. 
Additionally, we evaluate the transferability of the CAA steering vectors by applying them to two other BigToM tasks: \textit{Forward Action} and \textit{Backward Belief}. 
We provide details about these tasks in \Cref{app:bigtom}, and a more detailed explanation ITI in \Cref{app:iti}.

\section{Results}

\begin{figure*}[t]
  \centering
  \begin{subfigure}{\linewidth}
    \includegraphics[width=\linewidth]{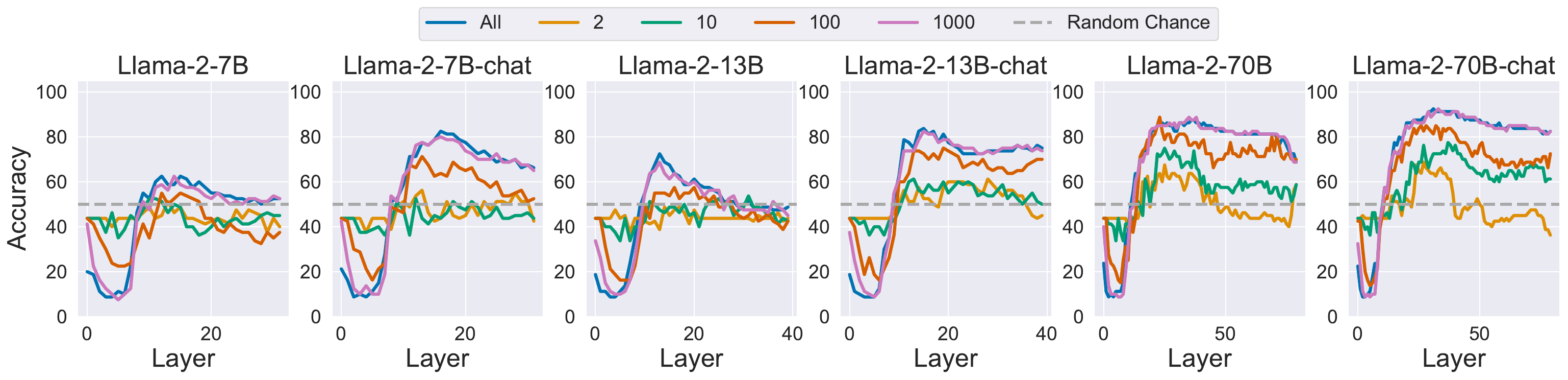}
  \end{subfigure}
  \caption{
  We compare the probing accuracy obtained by using the original set of activations (All) with the accuracy obtained by considering only the first $k=\{2, 10, 100, 1000\}$ principal components. 
  Results are for \textit{protagonist} beliefs (for \textit{oracle} see \Cref{fig:pca-oracle}). 
  In general, it is possible to recover most of the original accuracy by training probes on a smaller number $k$ of principal components of the activations.
  }
  \label{fig:pca}
\end{figure*}

\begin{figure*}[t]
  \centering
  \begin{subfigure}{\linewidth}
    \includegraphics[width=\linewidth]{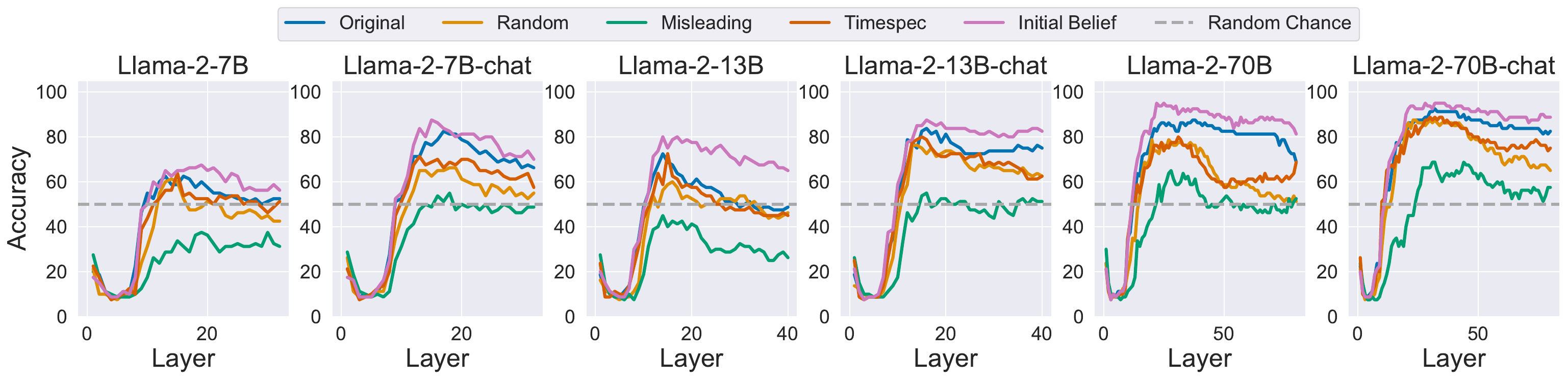}
  \end{subfigure}
  \caption{Sensitivity of \textit{protagonist} belief probing accuracy to different prompt variations. 
  Results for Pythia are shown in \Cref{fig:prompt-pythia}.
  Representations are brittle to prompt variations.
  }
  \label{fig:prompt}
\end{figure*}

\paragraph{Effect of Model Size and Fine-tuning}
Results from our study on model size and fine-tuning are shown in \Cref{fig:size-ft}. 
For \textit{oracle} beliefs, probing accuracy rapidly converges to $100$, with larger models showing faster convergence. 
Even the smallest Pythia-70m achieves $95\%$ accuracy.  
For \textit{protagonist} beliefs, we notice a similar pattern across most models, where accuracy at early layers is particularly low and then increases at the intermediate layers.
What happens at early layers is overfitting, which may be caused by spurious features introduced by the initial coding strategy of language models, where individual token representations are mixed together \citep{gurnee2023finding}.
We further discuss this in \Cref{sec:overfitting}.

In general, probing accuracy increases with model size, although there is a performance gap between Llama-2 and Pythia. 
For example, Llama2-13B reaches around $80\%$ accuracy, while Pythia-12B achieves approximately $60\%$. 
This gap is likely due to Llama-2 being trained on nearly seven times more tokens than Pythia (cf.\ \Cref{tab:models_details}).
Probes from fine-tuned LMs show significantly better accuracy, with improvements of up to $+29\%$ for Llama2-7B-chat (SFT + RLHF) and $+26\%$ for Pythia-6.9B-chat (SFT) compared to probes from their base version. 
The same probes outperform (Llama-2) or are on par (Pythia) with probes trained on twice as large base models (12/13B).
This highlights a key role of fine-tuning in shaping belief representations in smaller LMs. 
The performance gap closes for the largest Llama2-70B, for which the improvements from fine-tuning are marginal.

We characterise the relationship between probe accuracy and model size in \Cref{fig:fits}, using the \textit{best} accuracy for each LM -- i.e., the highest accuracy among probes ${g_l}$ trained on activations ${a_l}$ for model $f$. For Llama-2 base and Pythia base, probing accuracy scales logarithmically with model size (\Cref{subfig:llama-base-max-acc-log}, \ref{subfig:pythia-base-max-acc-log}), while for fine-tuned Llama-2 models, it scales linearly (\Cref{subfig:llama-chat-max-acc}).

\paragraph{Control Tasks}
\Cref{fig:control-1} shows that probes trained on the control task consistently perform at random chance, confirming that higher probing accuracy in larger models meaningfully reflects a greater ability to extract ToM representations, rather than simply being a by-product of spurious correlations.
For Llama models, the probes generally exhibit selectivity: they achieve high accuracy when probing for beliefs but remain at chance level on control tasks. 
Pythia’s overall accuracy is too low to allow for selectivity.

\Cref{fig:pca} shows probing accuracy on \textit{protagonist} when training the probes on the top $k$ principal components of Llama-2's internal activations. 
We provide results for Pythia in \Cref{fig:pca-pythia}, and for all models on \textit{oracle} settings in \Cref{fig:pca-oracle}. 
We consider $k=\{2, 10, 100, 1000\}$, spanning several orders of magnitude.\footnote{For models with hidden dimensions smaller than $1000$, we skip this value.}
Results show that it is generally possible to recover most of the original accuracy by training probes on a smaller number $k$ of principal components of the activations. 
We also performed the first control experiment, this time only using the first $k=\{100, 1000\}$ principal components. 
\Cref{fig:control-1000} and \ref{fig:control-100} again show that probes trained on the control task consistently perform at random chance, confirming that probes are not fitting spurious patterns.
Additionally, this suggests that belief representations are embedded in a low-dimensional subspace $\mathcal{B}$ spanned by the top $k$ eigenvectors $\{v_1,\dots, v_k\}$ of the covariance matrix $\mathsf{C} = \mathbb{E}[(a - \mathbb{E}[a])(a - \mathbb{E}[a])^\top]$.

\paragraph{Sensitivity to Prompting}
\Cref{fig:prompt} compares \textit{protagonist} probe accuracy across various prompt variations for Llama-2 models. 
As can be seen from the figure, providing the protagonist's \textit{Initial Belief} in the story yields higher probe accuracy compared to the \textit{Original} prompt.
Accuracy for all the other prompt variations is generally lower than \textit{Original}. 
\textit{Misleading} prompts hurt performance across all models.
This finding resonates with \citet{webson2022prompt}, who found that instruction-tuned models, despite being more robust, are still sensitive to misleading prompts. 
On the other hand, \textit{Time Specification} unexpectedly does not help in disambiguating belief states in different time frames, as we hypothesised in \S\ref{subs:probing-exp}. 
Additionally, models show sensitivity to \textit{Random} tokens placed before the belief statement. 
Pythia models show similar patterns, shown in \Cref{fig:prompt-pythia}.
Results for \textit{oracle} beliefs are reported in \Cref{fig:prompt-oracle} and indicate that models maintain high accuracy. \textit{Misleading} prompts slightly reduce performance to around $95\%$. 
In summary, these experiments show that LMs possess robust belief representations when taking an omniscient perspective, whereas their representations of others' beliefs are brittle to prompt variations.

\begin{table*}[t]
    \centering
    \small{
    \begin{tabular}{lllllllllll} 
        \toprule
        \multirow{2}{*}{\textbf{Model}} & \multirow{2}{*}{\textbf{Method}} & \multicolumn{3}{c}{\textbf{Forward Belief}} & \multicolumn{3}{c}{\textbf{Forward Action}} & \multicolumn{3}{c}{\textbf{Backward Belief}} \\
        \cmidrule(lr){3-5} \cmidrule(lr){6-8} \cmidrule(lr){9-11} 
        & & TB & FB & Both & TB & FB & Both & TB & FB & Both \\ 
        \midrule

        \multirow[t]{3}{*}{Llama-2-7b} & No int. & $44$ & $44$ & $44$ & $44$ & $44$ & $44$ & $44$ & $44$ & $44$ \\
        & ITI & $44_{+0}$ & $44_{+0}$ & $44_{+0}$ & $54_{\textcolor{PineGreen}{+10}}$ & $54_{\textcolor{PineGreen}{+10}}$ & $54_{\textcolor{PineGreen}{+10}}$ & $54_{\textcolor{PineGreen}{+10}}$ & $54_{\textcolor{PineGreen}{+10}}$ & $54_{\textcolor{PineGreen}{+10}}$ \\
        \rowcolor{gray!20}
        \cellcolor{white}
        & CAA & $66_{\textcolor{PineGreen}{+22}}^*$ & $71_{\textcolor{PineGreen}{+27}}^*$ & $54_{\textcolor{PineGreen}{+10}}$ & $66_{\textcolor{PineGreen}{+22}}^*$ & $57_{\textcolor{PineGreen}{+13}}^*$ & $54_{\textcolor{PineGreen}{+10}}$ & $60_{\textcolor{PineGreen}{+16}}^*$ & $74_{\textcolor{PineGreen}{+30}}$ & $54_{\textcolor{PineGreen}{+10}}$ \\

        \multirow[t]{3}{*}{Llama-2-7b-chat} & No int. & $56$ & $56$ & $55$ & $69$ & $55$ & $37$ & $56$ & $56$ & $55$ \\
        & ITI & $58_{\textcolor{PineGreen}{+2}}$ & $58_{\textcolor{PineGreen}{+2}}$ & $57_{\textcolor{PineGreen}{+2}}$ & $69_{+0}$ & $55_{+0}$ & $37_{+0}$ & $58_{\textcolor{PineGreen}{+2}}$ & $60_{\textcolor{PineGreen}{+3}}$ & $57_{\textcolor{PineGreen}{+2}}$ \\
        \rowcolor{gray!20}
        \cellcolor{white}
        & CAA & $70_{\textcolor{PineGreen}{+14}}$ & $72_{\textcolor{PineGreen}{+16}}^*$ & $57_{\textcolor{PineGreen}{+2}}$ & $69_{+0}$ & $67_{\textcolor{PineGreen}{+12}}$ & $53_{\textcolor{PineGreen}{+16}}$ & $66_{\textcolor{PineGreen}{+10}}$ & $84_{\textcolor{PineGreen}{+27}}^*$ & $57_{\textcolor{PineGreen}{+2}}^*$ \\

        \multirow[t]{3}{*}{Llama-2-13b} & No int. & $52$ & $44$ & $35$ & $59$ & $50$ & $37$ & $46$ & $49$ & $33$ \\
        & ITI & $52_{+0}$ & $45_{\textcolor{PineGreen}{+1}}$ & $35_{+0}$ & $64_{\textcolor{PineGreen}{+5}}$ & $61_{\textcolor{PineGreen}{+11}}$ & $46_{\textcolor{PineGreen}{+9}}$ & $48_{\textcolor{PineGreen}{+2}}$ & $59_{\textcolor{PineGreen}{+10}}$ & $42_{\textcolor{PineGreen}{+9}}$ \\
        \rowcolor{gray!20}
        \cellcolor{white}
        & CAA & $85_{\textcolor{PineGreen}{+33}}^*$ & $88_{\textcolor{PineGreen}{+44}}^*$ & $66_{\textcolor{PineGreen}{+31}}^*$ & $71_{\textcolor{PineGreen}{+12}}^*$ & $69_{\textcolor{PineGreen}{+19}}^*$ & $55_{\textcolor{PineGreen}{+18}}^*$ & $75_{\textcolor{PineGreen}{+29}}^*$ & $92_{\textcolor{PineGreen}{+43}}^*$ & $59_{\textcolor{PineGreen}{+26}}^*$ \\

        \multirow[t]{3}{*}{Llama-2-13b-chat} & No int. & $84$ & $56$ & $47$ & $78$ & $51$ & $38$ & $72$ & $48$ & $31$ \\
        & ITI & $84_{+0}$ & $65_{\textcolor{PineGreen}{+9}}$ & $59_{\textcolor{PineGreen}{+12}}$ & $78_{+0}$ & $58_{\textcolor{PineGreen}{+7}}$ & $47_{\textcolor{PineGreen}{+9}}^*$ & $72_{+0}$ & $60_{\textcolor{PineGreen}{+12}}$ & $48_{\textcolor{PineGreen}{+17}}$ \\
        \rowcolor{gray!20}
        \cellcolor{white}
        & CAA & $97_{\textcolor{PineGreen}{+13}}^*$ & $94_{\textcolor{PineGreen}{+38}}^*$ & $91_{\textcolor{PineGreen}{+44}}^*$ & $80_{\textcolor{PineGreen}{+2}}^*$ & $71_{\textcolor{PineGreen}{+20}}^*$ & $54_{\textcolor{PineGreen}{+16}}^*$ & $97_{\textcolor{PineGreen}{+25}}$ & $94_{\textcolor{PineGreen}{+46}}^*$ & $87_{\textcolor{PineGreen}{+56}}^*$ \\

        \multirow[t]{3}{*}{Llama-2-70b} & No int. & $90$ & $87$ & $78$ & $93$ & $52$ & $48$ & $73$ & $53$ & $32$ \\
        & ITI & $90_{+0}$ & $90_{\textcolor{PineGreen}{+3}}$ & $78_{+0}$ & $94_{\textcolor{PineGreen}{+1}}$ & $55_{\textcolor{PineGreen}{+3}}$ & $50_{\textcolor{PineGreen}{+2}}$ & $77_{\textcolor{PineGreen}{+4}}$ & $58_{\textcolor{PineGreen}{+5}}$ & $37_{\textcolor{PineGreen}{+5}}$ \\
        \rowcolor{gray!20}
        \cellcolor{white}
        & CAA & $99_{\textcolor{PineGreen}{+9}}^*$ & $97_{\textcolor{PineGreen}{+10}}^*$ & $95_{\textcolor{PineGreen}{+17}}^*$ & $94_{\textcolor{PineGreen}{+1}}^*$ & $80_{\textcolor{PineGreen}{+28}}^*$ & $73_{\textcolor{PineGreen}{+25}}^*$ & $94_{\textcolor{PineGreen}{+21}}$ & $92_{\textcolor{PineGreen}{+39}}^*$ & $83_{\textcolor{PineGreen}{+51}}^*$ \\

        \multirow[t]{3}{*}{Llama-2-70b-chat} & No int. & $69$ & $75$ & $56$ & $86$ & $56$ & $52$ & $63$ & $59$ & $52$ \\
        & ITI & $69_{+0}$ & $76_{\textcolor{PineGreen}{+1}}$ & $59_{\textcolor{PineGreen}{+2}}$ & $86_{+0}$ & $56_{+0}$ & $52_{+0}$ & $63_{+0}$ & $60_{\textcolor{PineGreen}{+1}}$ & $54_{\textcolor{PineGreen}{+2}}$ \\
        \rowcolor{gray!20}
        \cellcolor{white}
        & CAA & $92_{\textcolor{PineGreen}{+23}}^*$ & $97_{\textcolor{PineGreen}{+22}}^*$ & $89_{\textcolor{PineGreen}{+32}}^*$ & $87_{\textcolor{PineGreen}{+1}}^*$ & $75_{\textcolor{PineGreen}{+19}}^*$ & $60_{\textcolor{PineGreen}{+8}}^*$ & $88_{\textcolor{PineGreen}{+25}}$ & $92_{\textcolor{PineGreen}{+33}}^*$ & $80_{\textcolor{PineGreen}{+28}}$ \\
        
        \bottomrule
    \end{tabular}
    }
    \caption{Comparison of the effects of ITI \citep{li2023inferencetime} and CAA \citep{rimsky2023steering} on three tasks from BigToM \citep{gandhi2024understanding}. TB denotes a true belief task, whereas FB denotes a false belief task. The numbers represent accuracy scores, with the difference in performance compared to no intervention (No int.) indicated as subscripts. 
    The asterisk ($*$) denotes a statistically significant difference from No int.\ based on a t-test with $p < 0.05$. Results for Pythia are shown in \Cref{tab:caa_params}.
    CAA outperforms ITI on all tasks.
    }
    \label{tab:caa_results}
\end{table*}

\paragraph{Contrastive Activation Addition}
We compare models' accuracy on three BigToM tasks in \Cref{tab:caa_results} (Llama) and \Cref{tab:caa_params} (Pythia).
Each model has been evaluated three times: without any intervention, using ITI, and using CAA. 
Hyperparameter details can be found in \Cref{app:caa}.
Note that we use steering vectors computed using the \textit{Forward Belief} task for \underline{all} three tasks to test their generalisability.

Performance without intervention is generally lower across tasks and model sizes, with the larger Llama-2-70B and Llama-2-70B-chat models exhibiting higher accuracy. 
Performance for Pythia models of different sizes does not change much, with the fine-tuned Pythia-6.9B-chat often showing better performance on single true belief (TB) and false belief (FB) tasks but not on their conjunction (Both). 

ITI demonstrates modest improvements over no intervention for Llama-2 models.
Improvements for Pythia models are consistent and higher, up to $+17$. 
The only exception is Pythia-6.9B-chat, for which ITI is not always beneficial. 

CAA consistently delivers the most substantial accuracy improvements across all models and tasks, up to $+56$ for Llama-2-13B-chat on the \textit{Backward Belief} task, which \citeauthor{gandhi2024understanding} have identified as the hardest task. 
Despite its relatively small size, Llama-2-13B-chat excels in all three tasks when using CAA.
Larger 70B models often achieve accuracies close to or exceeding $90\%$. 
Smaller models like Pythia-70M and Pythia-410M also show significant gains with CAA, though the absolute performance is still lower than Llama-2.
To further demonstrate CAA's effectiveness, we applied it while evaluating models on a control task where the causal event in the story is replaced by a random one that does not change the environment (e.g., \textit{A musician starts playing music while Noor is making the latte}). 
\Cref{tab:caa_control} shows improved results for all models, indicating that CAA improves performance on ToM tasks without compromising the models' ability on control tasks.

Overall, our results indicate that it is possible to further enhance ToM reasoning in LMs in a computationally cheap way, without needing to train any probe.
Furthermore, we show that the CAA steering vectors are general, yielding substantial performance gains across all ToM tasks. 

\section{Discussion and Conclusion}
\vspace{-0.1cm}
In this work, we conducted extensive experiments across 12 LMs to examine their internal representation of beliefs of self (\textit{oracle}) and others (\textit{protagonist}).
Our experiments show \textbf{similar emergence patterns across all the models we evaluated (RQ1)}: \textit{oracle} belief representations generally form in the first layers, while for \textit{protagonist} they emerge at the intermediate layers. 
Moreover, \textbf{probing accuracy increases with model size and, more crucially for smaller models, with fine-tuning (RQ1)} (\Cref{fig:size-ft}).
While larger models show higher probing accuracy, this could be due to their higher dimensionality -- at the same time increasing the number of learning parameters in the probes and offering more spurious patterns to fit. 
To control for this, we ran two experiments: one using randomly permuted labels, and one projecting activations onto their top-$k$ principal components to reduce probe size. 
Results show that high-dimensional probes cannot learn random label mappings (Fig.~\ref{fig:control}), and that reduced representations retain most of the original accuracy (Fig.~\ref{fig:pca}, \ref{fig:pca-pythia}, \ref{fig:pca-oracle}). 
Together, these findings suggest that \textbf{probes capture structured belief representations rather than spurious correlations (RQ2)}.
We then explore if these representations are robust to prompt variations.
Our experiments demonstrate that \textbf{LMs possess robust belief representations when taking an omniscient perspective} (Fig.~\ref{fig:prompt-oracle}), \textbf{whereas their representations of others' beliefs are more brittle (RQ3)}, with probing accuracy decreasing for semantically neutral prompts (Fig.~\ref{fig:prompt}, \ref{fig:prompt-pythia}).  
Our final set of experiments shows that \textbf{belief representations can be strengthened using CAA (RQ4)}. 
CAA steers model activations in a generalisable way, significantly improving performance across multiple ToM tasks while being computationally cheaper than ITI (\Cref{tab:caa_results}, \ref{tab:caa_params}). 
For instance, with Llama-2-70B, ITI requires training 5,120 probes 
(64 attention heads $\times$ 80 layers), whereas CAA only needs 80 vectors, one per layer.

In summary, our key takeaway is that while models can robustly represent beliefs from an omniscient perspective, 
\begin{quote}
    \textit{representations of others' beliefs improve with model size and fine-tuning, are structured yet brittle -- but also easily steerable.}
\end{quote}

Together, our findings suggest several promising directions for future work. 
Better understanding the similar emergence pattern of belief representations across LMs can inform architecture design and training strategies.
Especially for smaller models, future work could explore how different types of fine-tuning (e.g., human feedback vs. synthetic data) influence the emergence of internal belief representations.
Demonstrating that these representations are structured rather than spurious validates the use of probing as a meaningful tool to study how LMs' represent beliefs of self and others, and encourages internal model analysis as part of evaluation pipelines. 
However, the brittleness of belief representations to prompts -- particularly when attributing beliefs to others -- suggests that the perspective-taking machinery needed for robust ToM reasoning remains fragile, and highlights the need for robustness benchmarks and new approaches to improve generalisation. 
Finally, our success with CAA shows that belief representations can be strengthened in a generalisable and efficient way, opening up opportunities for real-time model steering in socially grounded tasks.
While CAA offers a post-hoc remedy, future research should also explore methods for directly embedding perspective-taking circuits into model architectures.

\section*{Limitations}

Our study focused on expanding experiments from the model perspective, examining architectures, sizes, fine-tuning, and prompt design, all within the same dataset. 
A natural extension of our work is replicating these experiments across multiple datasets and more model families. 
Given the rapid pace of new language model releases, studying all available models is impractical, particularly considering computational resource constraints. 
Nevertheless, our approach can be adopted to support new benchmarks or to evaluate newly released models as they become available.
Finally, while in this work we focused on beliefs, our experimental approach can be adapted to investigate how LMs represent desires, emotions, intentions, or preferences. 
Future research exploring other types of mental states can use our findings to determine whether similar or distinct patterns emerge.

\section*{Acknowledgements}
L. Shi was funded by the Deutsche Forschungsgemeinschaft (DFG, German Research Foundation) under Germany's Excellence Strategy -- EXC 2075 -- 390740016.
The authors thank the International Max Planck Research School for Intelligent Systems (IMPRS-IS) for supporting C. Ruhdorfer.
\bibliography{tom_ref,refs}
\appendix
\section{Appendix}

\subsection{Experimental setup}
\label{app:setup}

\subsubsection{BigToM} 
\label{app:bigtom}
BigToM \citep{gandhi2024understanding} is constructed using GPT-4 \citep{achiam2023gpt} to populate causal templates and combine elements from these templates. 
Each causal template is set up with a \textit{context} and a description of the \textit{protagonist} (e.g. \textit{``Noor is working as a barista [\dots]''}), a \textit{desire} (\textit{``Noor wants to make a cappuccino''}), a \textit{percept} (\textit{``Noor grabs a milk pitcher and fills it with oat milk''}), and a \textit{belief} (\textit{``Noor believes that the pitcher contains oat milk''}). 
The state of the world is changed by a \textit{causal event} (\textit{``A coworker swaps the oat milk in the pitcher with almond milk''}).
The dataset constructs different conditions by changing the percepts of the protagonist after the causal event, which will result in different beliefs -- true or false.
\citet{gandhi2024understanding} generated 200 templates and extracted 25 conditions from each template, resulting in 5,000 test samples. 
In this work, following \citet{zhu2024language} and \citet{gandhi2024understanding} we focused on the 6 most important conditions, corresponding to true and false beliefs on the following three tasks: 
\begin{itemize}
    \item \textit{Forward Belief}: given the protagonist’s percepts of the causal event, infer their belief: $P(\mathrm{belief} | \mathrm{percept})$.
    \item \textit{Forward Action}: infer the protagonist’s action given their desire and percepts of the causal event. Before inferring the action, one would need to first implicitly infer the protagonist's belief: $\sum_{\mathrm{belief}} P(\mathrm{action} | \mathrm{percept}, \mathrm{belief}, \mathrm{desire})$.
    \item \textit{Backward Belief}: infer the protagonist’s belief from observed actions. This requires to first implicitly infer the protagonist's percepts: $\sum_{\mathrm{percepts}} P(\mathrm{belief} | \mathrm{action}, \mathrm{percept}, \mathrm{desire})$.
\end{itemize} 
The dataset was released under the MIT license and can be accessed at \url{https://github.com/cicl-stanford/procedural-evals-tom}. 
We report one example for each task in Example~\ref{box:fb}, \ref{box:fa}, and \ref{box:bb}, where the text defining true belief or false belief task is shown in \textcolor{ablue}{blue} and \textcolor{ared}{red}, respectively. 

\subsubsection{Linear probes} 
\label{app:probes}
\begin{figure*}[t]
    \centering
    \includegraphics[width=\textwidth]{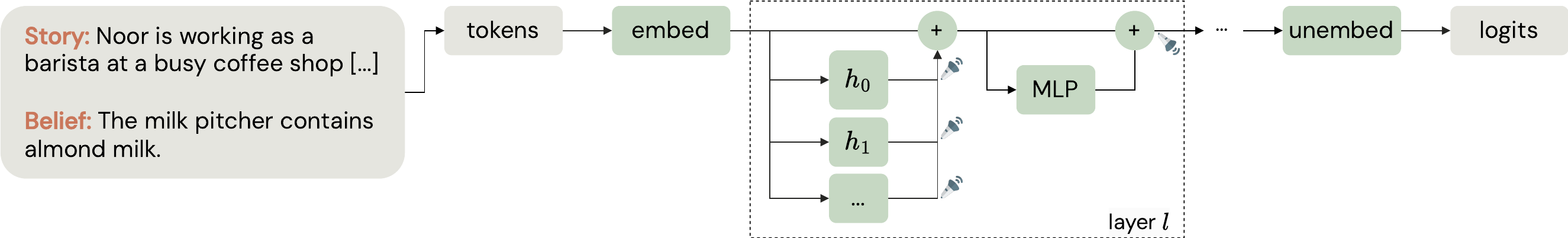}
    \caption{Given a tokenised input, we cache the internal activations for all attention heads $h_i$, $i=0, \dots, H-1$, and residual streams. In our experiments, we use residual stream activations.}
    \label{fig:probe}
\end{figure*}

Our probing approach is illustrated in \Cref{fig:probe}. 
For our experiments, we cache activations at the residual stream level. 
To perform ITI and compare it to CAA, we also cache attention heads activations. 
We trained the probes using the L-BFGS solver \citep{liu1989limited} with L2 penalty with inverse of regularisation strength $10$ for a maximum of $1000$ iterations. We use zero as random seed. 

\subsubsection{Language models} 
\label{app:lms}

A detailed summary of the models we use in this work is shown in \Cref{tab:models_details}. 
Pythia was released under the Apache 2.0 license.
Llama-2 is licensed by Meta for both researchers and commercial entities \citep{touvron2023llama}. 
For all the models, we set the temperature to zero.

\begin{table*}[t]
    \centering
    \begin{tabular}{ccccccc}
        \toprule
        LM  & Size & + SFT & + RLHF & Tokens & $d_{model}$ & Layers \\
        \midrule
        \multirow{3}{*}{Llama-2} & 7B & & & 2T & 4096 & 32 \\
        & 13B & & & 2T & 5120 & 40 \\
        & 70B & & & 2T & 8192 & 80 \\
        \multirow{3}{*}{Llama-2-chat} & 7B & \checkmark & \checkmark & 2T & 4096 & 32 \\
        & 13B & \checkmark & \checkmark & 2T & 5120 & 40 \\
        & 70B & \checkmark & \checkmark & 2T & 8192 & 80 \\
        \midrule
        \multirow{6}{*}{Pythia} & 70M & & & 300B & 512 & 6 \\
        & 410M & & & 300B & 1024 & 24 \\
        & 1B & & & 300B & 2048 & 16 \\
        & 6.9B & & & 300B & 4096 & 32 \\
        & 12B & & & 300B & 5120 & 36 \\
        & 6.9B & \checkmark & & 300B & 4096 & 32 \\
        \bottomrule
    \end{tabular}
    \caption{The 12 models used in this work.}
    \label{tab:models_details}
\end{table*} 

\subsubsection{Examples of prompt variations}
\label{app:prompts}
Example~\ref{box:default} shows an example of \textit{Original} prompt. 
Examples of prompt variations are provided in Example~\ref{box:random} (\textit{Random}), Example~\ref{box:misleading} (\textit{Misleading}), Example~\ref{box:time} (\textit{Time Specification}), and Example~\ref{box:init} (\textit{Initial Belief}).

\subsection{Model size and fine-tuning}
\label{app:fits}
\begin{figure*}[t]
  \centering
  \begin{subfigure}{0.23\linewidth}
    \includegraphics[width=\linewidth]{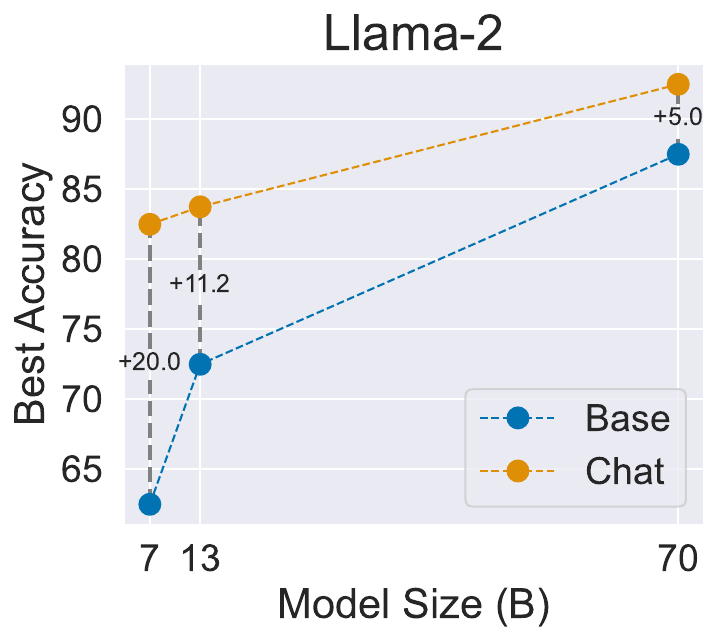}
    \caption{}
    \label{subfig:llama-max-acc}
  \end{subfigure}
  \begin{subfigure}{0.23\linewidth}
    \includegraphics[width=\linewidth]{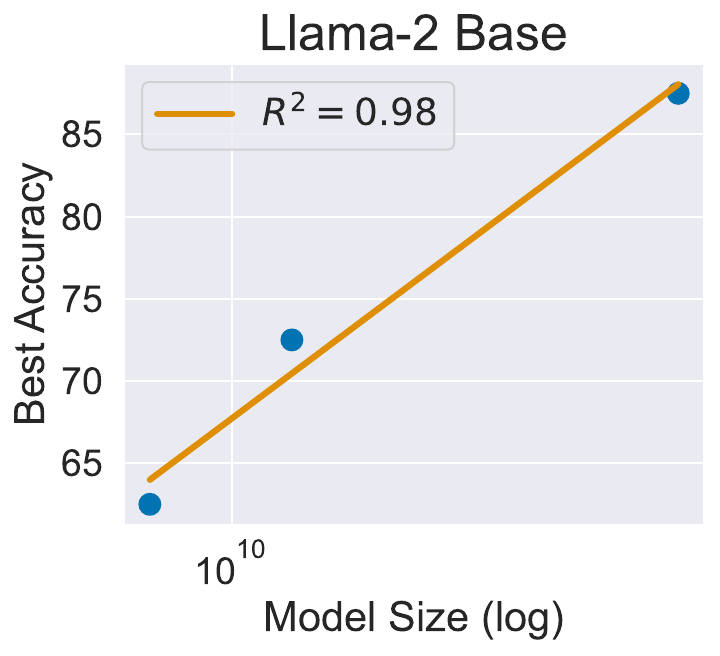}
    \caption{}
    \label{subfig:llama-base-max-acc-log}
  \end{subfigure}
  \begin{subfigure}{0.23\linewidth}
    \includegraphics[width=\linewidth]{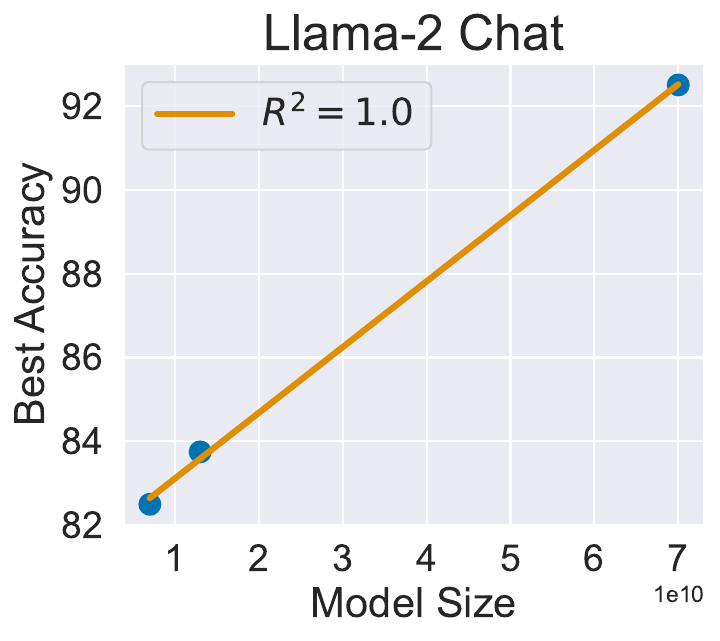}
    \caption{}
    \label{subfig:llama-chat-max-acc}
  \end{subfigure}
  \begin{subfigure}{0.23\linewidth}
    \includegraphics[width=\linewidth]{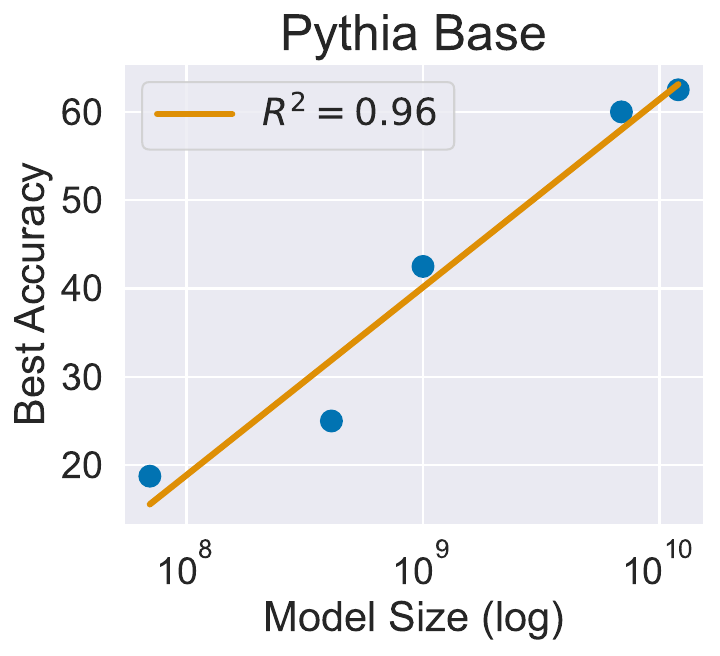}
    \caption{}
    \label{subfig:pythia-base-max-acc-log}
  \end{subfigure}
  \caption{To characterise the relationship between probe accuracy and model size we consider the \textit{best} probe accuracy for every LM, i.e. the highest accuracy among probes $\{g_l\}$ trained on $\{a_l\}$ for a LM $f$. 
  \textbf{(a)} Best accuracy for Llama-2 models of different size. Numbers on the vertical dotted lines indicate the gain in accuracy between base and fine-tuned model of the same size. \textbf{(b)} Logarithmic fit for Llama-2 base. \textbf{(c)} Linear fit for Llama-2 fine-tuned (chat). \textbf{(d)} Logarithmic fit for Pythia base.}
  \label{fig:fits}
\end{figure*}

To characterise the relationship between probe accuracy and model size we consider the \textit{best} probe accuracy for every LM, i.e. the highest accuracy among probes $\{g_l\}$ trained on $\{a_l\}$ for a LM $f$.
For Llama-2 base, the best probe accuracy scales logarithmically with model size ($R^2 = 0.98$, \Cref{subfig:llama-base-max-acc-log}), whereas for fine-tuned models it scales linearly ($R=1.0$, cf.\ \Cref{subfig:llama-chat-max-acc}).
For Pythia base, the best probe accuracy also scales logarithmically with model size ($R^2 = 0.96$, \Cref{subfig:pythia-base-max-acc-log}).  

\subsubsection{Overfitting Issues}
\label{sec:overfitting}
\Cref{fig:size-ft} also that probing accuracy at early layers is particularly low across all models, performing even worse than random. 
This happens due to overfitting, which may be caused by spurious features introduced by the initial coding strategy of language models, where individual token representations are mixed together \citep{gurnee2023finding}.
We also identified the same issue when reproducing the results in \citet{zhu2024language}, who address it by manually clipping all accuracies below random chance to 50\%.\footnote{\url{https://github.com/Walter0807/RepBelief/blob/0fc86396f2f0a998643ea01786eb3db4dd20ff9c/probe.py\#L60}}
Since probing experiments require training a large number of probes for each model, both we and \citet{zhu2024language} trained each probe for the same fixed number of epochs (1,000). 
However, for activations from the earlier layers, overfitting occurs very quickly - often within the first 10 iterations. 

We ran an experiment with Llama2-7B-chat, reducing training to fewer than 10 iterations, and found that the probes performed at random chance.
Therefore, to fully resolve this issue, we would need to choose the number of training epochs for each probe individually. 
This would likely flatten the observed "U" shape in the results. 
However, this process would be computationally expensive and does not contribute to our main research questions.
Rather than artificially adjusting accuracies to 50\%, we prefer to present the results as they are.

\begin{figure*}
    \centering
    \begin{subfigure}{\linewidth}
        \includegraphics[width=\linewidth]{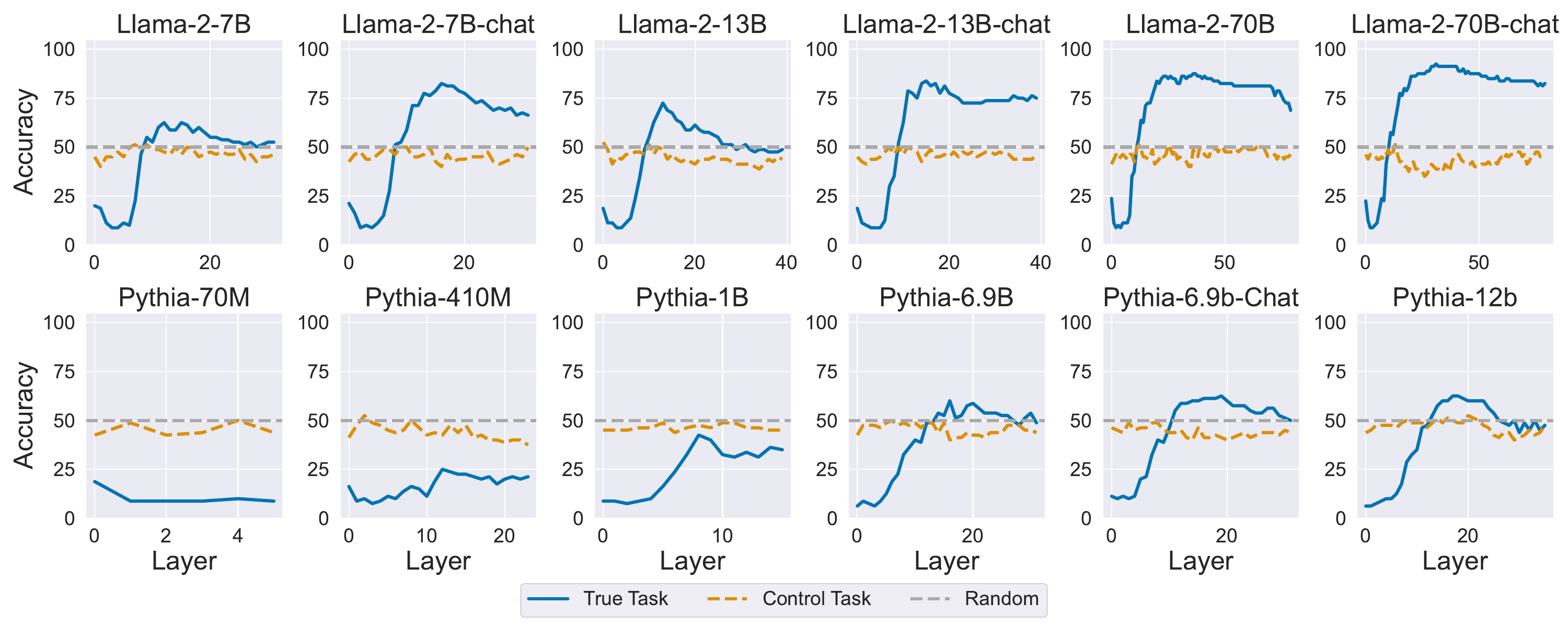}
        \caption{Original activations.}
        \label{fig:control-1}
    \end{subfigure}
    \begin{subfigure}{\linewidth}
        \includegraphics[width=\linewidth]{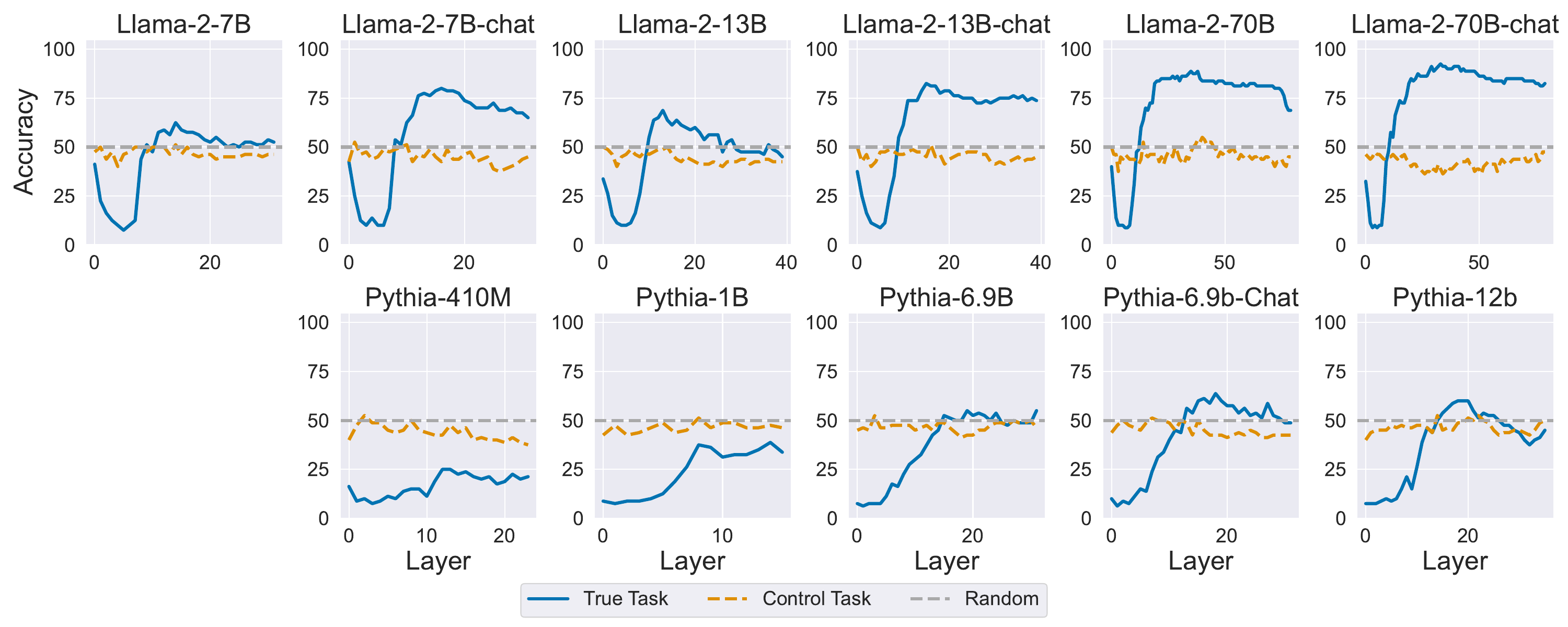}
        \caption{$k=1000$ largest principal components of the activations.}
        \label{fig:control-1000}
    \end{subfigure}
    \begin{subfigure}{\linewidth}
        \includegraphics[width=\linewidth]{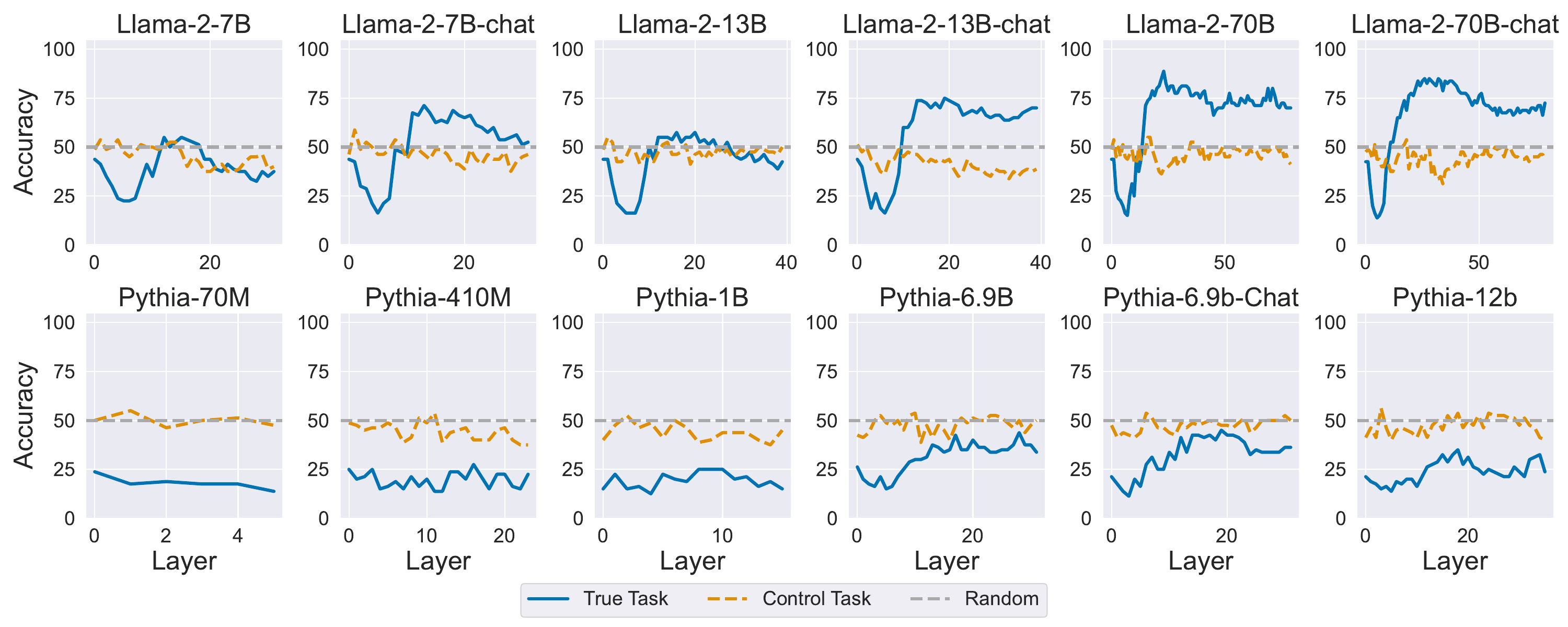}
        \caption{$k=100$ largest principal components of the activations.}
        \label{fig:control-100}
    \end{subfigure}
    \caption{Comparison between accuracy on belief probing and accuracy obtained on a control task.}
    \label{fig:control}
\end{figure*}

\subsection{Sensitivity to prompting}
\label{app:sensitivity-to-prompting}
\begin{figure*}[t]
  \centering
  \begin{subfigure}{\linewidth}
    \includegraphics[width=\linewidth]{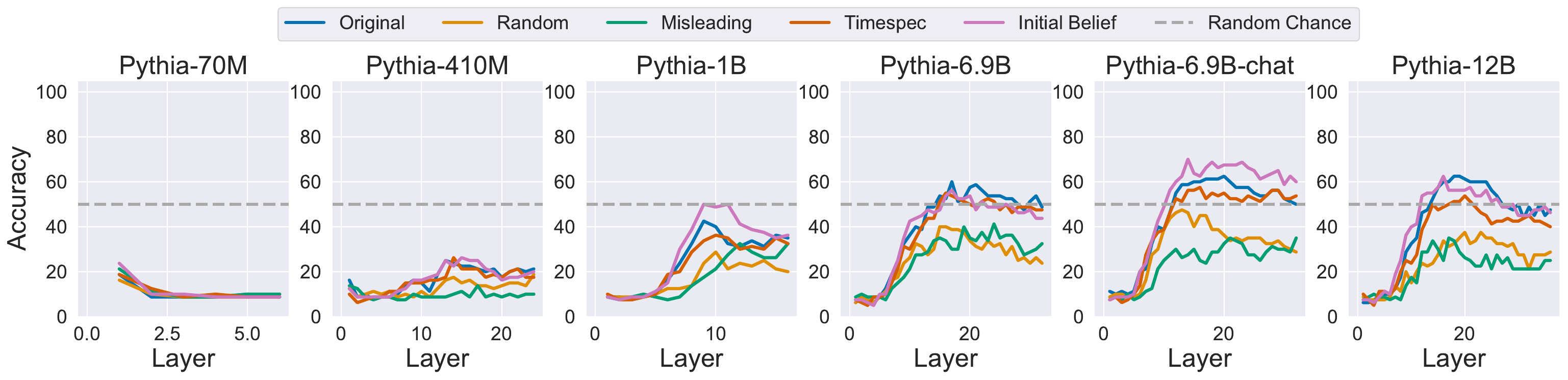}
  \end{subfigure}
  \caption{Sensitivity of protagonist belief probing accuracy to different prompt variations.}
  \label{fig:prompt-pythia}
\end{figure*}

Accuracy on \textit{protagonist} belief probing for Pythia models is shown in \Cref{fig:prompt-pythia}.

\begin{figure*}[t]
  \centering
  \begin{subfigure}{\linewidth}
    \includegraphics[width=\linewidth]{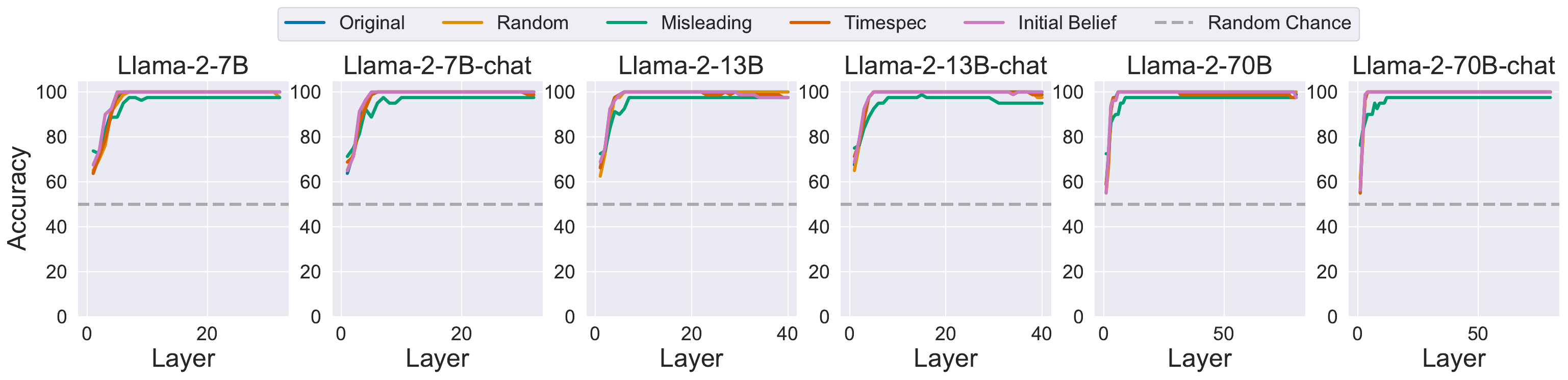}
  \end{subfigure}
  \begin{subfigure}{\linewidth}
    \includegraphics[width=\linewidth]{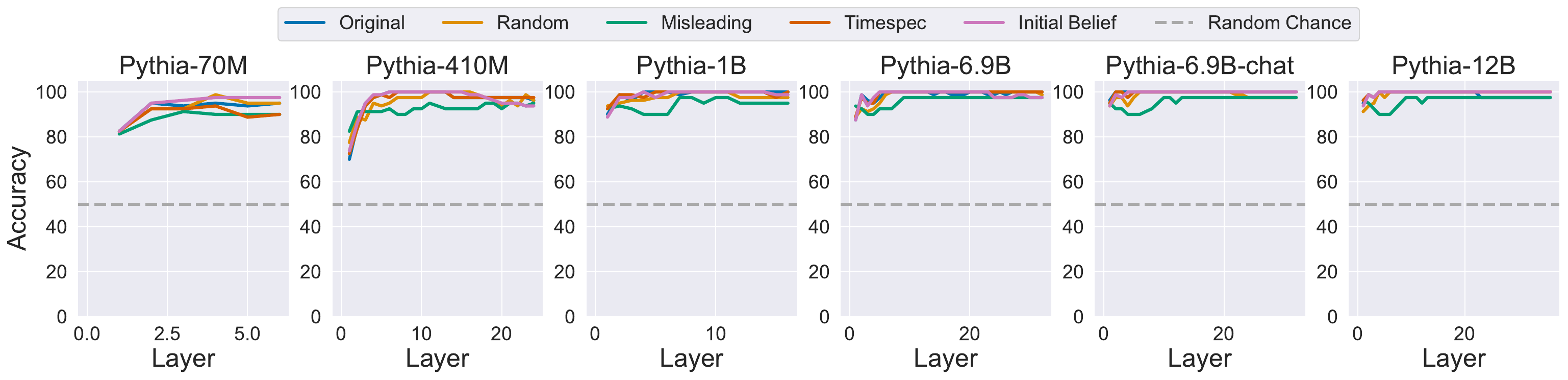}
  \end{subfigure}
  \caption{Sensitivity of protagonist belief probing accuracy to different prompt variations.}
  \label{fig:prompt-oracle}
\end{figure*}

Accuracy on \textit{oracle} belief probing for different prompt variations are reported in \Cref{fig:prompt-oracle}.

\subsection{Dimensionality reduction}
\begin{figure*}[t]
  \centering
  \begin{subfigure}{\linewidth}
    \includegraphics[width=\linewidth]{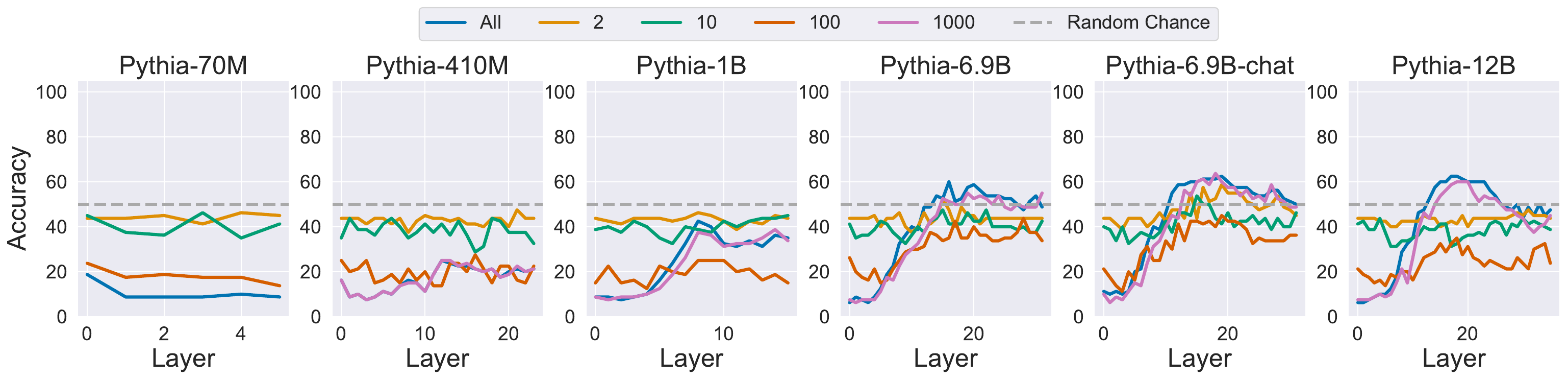}
  \end{subfigure}
  \caption{
  We compare the probing accuracy obtained by using the original set of activations (All) with the accuracy obtained by considering only the first $n=\{2, 10, 100, 1000\}$ principal components. 
  For Pythia: All(70m) = $512$, All(410m) = $1024$, All(1b) = $2048$, All(6.9b) = $4096$, All(12b) = $5120$. 
  Results for \textit{oracle} are shown in \Cref{fig:pca-oracle}.
  }
  \label{fig:pca-pythia}
\end{figure*}

Probing accuracy obtained by Pythia models for the \textit{protagonist} setting is reported in \Cref{fig:pca-pythia}.

\begin{figure*}[t]
  \centering
  \begin{subfigure}{\linewidth}
    \includegraphics[width=\linewidth]{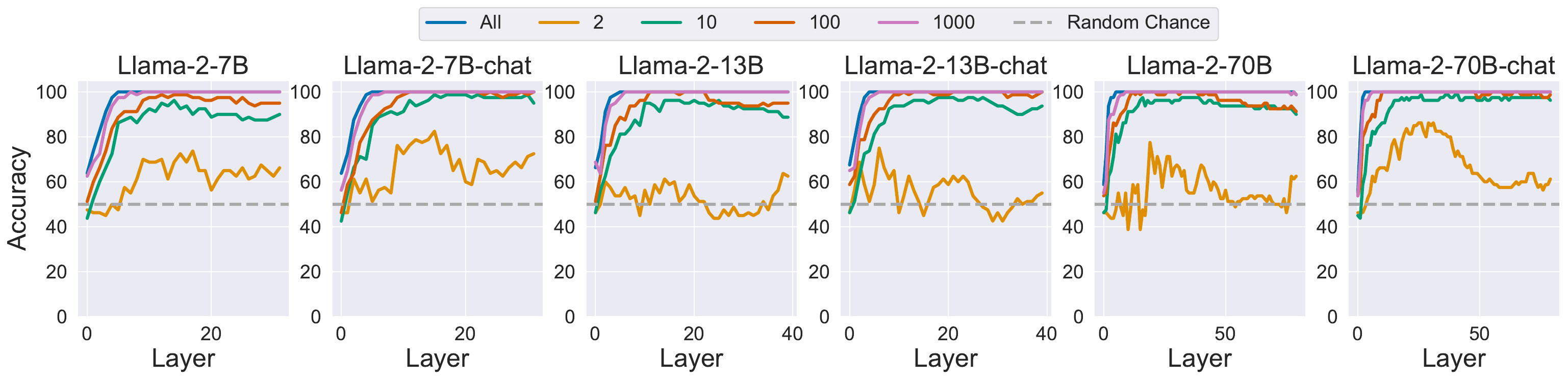}
  \end{subfigure}
  \begin{subfigure}{\linewidth}
    \includegraphics[width=\linewidth]{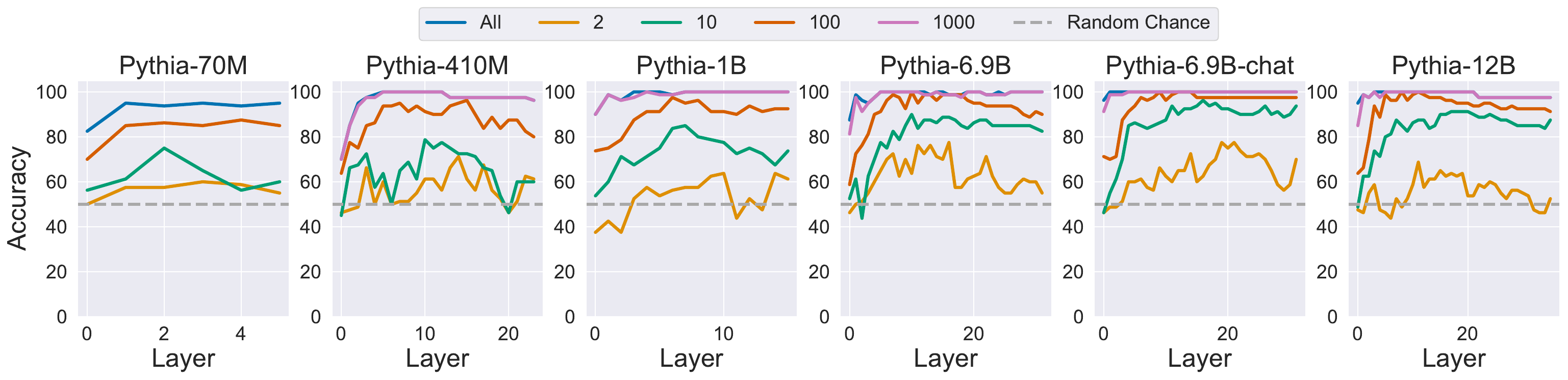}
  \end{subfigure}
  \caption{\textbf{(Oracle)} To investigate potential memorisation in the probes, we compare the probing accuracy obtained by using the original set of activations (All) with the accuracy obtained by considering only the first $n=\{2, 10, 100, 1000\}$ principal components. For Llama2: All(7b) = $4096$, All(13b) = $5120$, All(70b) = $8192$. For Pythia: All(70m) = $512$, All(410m) = $1024$, All(1b) = $2048$, All(6.9b) = $4096$, All(12b) = $5120$.}
  \label{fig:pca-oracle}
\end{figure*}

\textit{Oracle} probe accuracy obtained by considering only the first $n=\{2, 10, 100, 1000\}$ principal components are shown in \Cref{fig:pca-oracle}.

\subsection{Inference-time intervention}
\label{app:iti}
Inference-time intervention \citep[ITI]{li2023inferencetime} employs a two-step process. 
First, it trains a probe for each attention head across all layers of a LM. 
These probes are evaluated on a validation set, and the top-$k$ heads with the highest accuracy are selected.
Subsequently, during inference, ITI steers the activations of these top heads along the directions defined by their corresponding probes.
Formally, \textcolor{Plum}{ITI} can be defined as an additional term to the multi-head attention:
\begin{equation*}
    x_{l+1} = x_l + \sum_{h=1}^H Q_l^h \left( \mathrm{Att}_l^h (P_l^h x_l) \textcolor{Plum}{\ + \  \alpha\sigma_l^h\theta_l^h} \right)
\end{equation*}
where $x_l$ is the residual stream at layer $l$, $H$ is the number of attention heads, $\alpha \in \mathbb{R}^+$ is a coefficient, $\sigma_l^h$ is the standard deviation of activations along the direction identified by the probe trained on attention head $h$ at layer $l$, and $\theta_l^h$ is zero ofr not-selected attention heads.

\subsection{Activation editing}
\label{app:caa}

\Cref{tab:caa_params} reports results obtained on the three BigToM tasks with the corresponding hyperparameters used for ITI \citep{li2023inferencetime} and CAA \citep{rimsky2023steering}.
We report an example of prompt used for evaluation in Example~\ref{box:eval}.
\Cref{tab:caa_control} shows the accuracy obtained by using CAA on the Forward Belief True Control task in BigToM. 
On this control task, CAA produced improved results for all model, proving that CAA not only improves performance on ToM tasks, but also does not degrades the models' ability to perform other tasks.

\begin{table*}[t]
    \centering
    \resizebox{\linewidth}{!}{
    \begin{tabular}{llccccccccc}
        \toprule
        \multirow{2}{*}{\textbf{Model}} & \multirow{2}{*}{\textbf{Method}} & \multicolumn{3}{c}{\textbf{Forward Belief}} & \multicolumn{3}{c}{\textbf{Forward Action}} & \multicolumn{3}{c}{\textbf{Backward Belief}} \\
        \cmidrule(lr){3-5} \cmidrule(lr){6-8} \cmidrule(lr){9-11} 
        & & TB & FB & Both & TB & FB & Both & TB & FB & Both \\
        \midrule

        \multirow[t]{3}{*}{Llama-2-7b} & No int. & $44$ & $44$ & $44$ & $44$ & $44$ & $44$ & $44$ & $44$ & $44$ \\
        & ITI & $44_{0.0}$ & $44_{0.0}$ & $44_{0.0}$ & $54_{20.0}$ & $54_{20.0}$ & $54_{20.0}$ & $54_{20.0}$ & $54_{20.0}$ & $54_{20.0}$ \\
        \rowcolor{gray!20}
        \cellcolor{white}
        & CAA & $66_{2.0, 11}$ & $71_{1.0, 31}$ & $54_{2.0, 0}$ & $66_{2.0, 11}$ & $57_{2.0, 12}$ & $54_{2.0, 2}$ & $60_{2.0, 11}$ & $74_{1.0, 31}$ & $54_{2.0, 2}$ \\

        \multirow[t]{3}{*}{Llama-2-7b-chat} & No int. & $56$ & $56$ & $55$ & $69$ & $55$ & $37$ & $56$ & $56$ & $55$ \\
        & ITI & $58_{15.0}$ & $58_{15.0}$ & $57_{15.0}$ & $69_{0.0}$ & $55_{0.0}$ & $37_{0.0}$ & $58_{10.0}$ & $60_{10.0}$ & $57_{10.0}$ \\
        \rowcolor{gray!20}
        \cellcolor{white}
        & CAA & $70_{1.0, 11}$ & $72_{1.5, 10}$ & $57_{1.0, 1}$ & $69_{0.0, 0}$ & $67_{1.5, 11}$ & $53_{1.5, 12}$ & $66_{1.0, 11}$ & $84_{1.5, 10}$ & $57_{1.0, 0}$ \\

        \multirow[t]{3}{*}{Llama-2-13b} & No int. & $52$ & $44$ & $35$ & $59$ & $50$ & $37$ & $46$ & $49$ & $33$ \\
        & ITI & $52_{0.0}$ & $45_{15.0}$ & $35_{0.0}$ & $64_{15.0}$ & $61_{20.0}$ & $46_{20.0}$ & $48_{20.0}$ & $59_{20.0}$ & $42_{20.0}$ \\
        \rowcolor{gray!20}
        \cellcolor{white}
        & CAA & $85_{2.0, 12}$ & $88_{2.0, 14}$ & $66_{2.0, 12}$ & $71_{1.5, 10}$ & $69_{2.0, 13}$ & $55_{1.0, 39}$ & $75_{2.0, 10}$ & $92_{2.0, 13}$ & $59_{1.5, 12}$ \\

        \multirow[t]{3}{*}{Llama-2-13b-chat} & No int. & $84$ & $56$ & $47$ & $78$ & $51$ & $38$ & $72$ & $48$ & $31$ \\
        & ITI & $84_{0.0}$ & $65_{15.0}$ & $59_{15.0}$ & $78_{0.0}$ & $58_{15.0}$ & $47_{15.0}$ & $72_{0.0}$ & $60_{15.0}$ & $48_{15.0}$ \\
        \rowcolor{gray!20}
        \cellcolor{white}
        & CAA & $97_{1.0, 12}$ & $94_{1.0, 12}$ & $91_{1.0, 12}$ & $80_{1.5, 11}$ & $71_{1.0, 13}$ & $54_{1.5, 13}$ & $97_{1.5, 10}$ & $94_{1.5, 12}$ & $87_{1.5, 12}$ \\

        \multirow[t]{3}{*}{Llama-2-70b} & No int. & $90$ & $87$ & $78$ & $93$ & $52$ & $48$ & $73$ & $53$ & $32$ \\
        & ITI & $90_{0.0}$ & $90_{20.0}$ & $78_{0.0}$ & $94_{15.0}$ & $55_{20.0}$ & $50_{15.0}$ & $77_{10.0}$ & $58_{15.0}$ & $37_{10.0}$ \\
        \rowcolor{gray!20}
        \cellcolor{white}
        & CAA & $99_{2.0, 16}$ & $97_{1.5, 19}$ & $95_{1.5, 18}$ & $94_{1.5, 2}$ & $80_{2.0, 19}$ & $73_{1.5, 18}$ & $94_{2.0, 18}$ & $92_{2.0, 19}$ & $83_{1.5, 19}$ \\

        \multirow[t]{3}{*}{Llama-2-70b-chat} & No int. & $69$ & $75$ & $56$ & $86$ & $56$ & $52$ & $63$ & $59$ & $52$ \\
        & ITI & $69_{0.0}$ & $76_{10.0}$ & $59_{10.0}$ & $86_{0.0}$ & $56_{0.0}$ & $52_{0.0}$ & $63_{0.0}$ & $60_{10.0}$ & $54_{10.0}$ \\
        \rowcolor{gray!20}
        \cellcolor{white}
        & CAA & $92_{1.5, 18}$ & $97_{1.5, 25}$ & $89_{1.5, 18}$ & $87_{1.5, 17}$ & $75_{1.0, 19}$ & $60_{1.0, 19}$ & $88_{1.5, 18}$ & $92_{1.0, 19}$ & $80_{1.5, 18}$ \\

        \multirow[t]{3}{*}{Pythia-70m} & No int. & $41$ & $41$ & $37$ & $46$ & $45$ & $41$ & $44$ & $41$ & $37$ \\
        & ITI & $54_{20.0}$ & $54_{20.0}$ & $54_{20.0}$ & $54_{20.0}$ & $54_{20.0}$ & $54_{20.0}$ & $54_{20.0}$ & $54_{20.0}$ & $54_{20.0}$ \\
        \rowcolor{gray!20}
        \cellcolor{white}
        & CAA & $62_{1.0, 2}$ & $56_{1.0, 1}$ & $54_{1.5, 1}$ & $59_{1.0, 2}$ & $60_{1.0, 3}$ & $58_{1.0, 2}$ & $63_{1.0, 2}$ & $56_{1.0, 2}$ & $54_{1.5, 1}$ \\

        \multirow[t]{3}{*}{Pythia-410m} & No int. & $48$ & $45$ & $45$ & $44$ & $44$ & $44$ & $44$ & $47$ & $44$ \\
        & ITI & $55_{20.0}$ & $62_{20.0}$ & $52_{20.0}$ & $54_{20.0}$ & $54_{20.0}$ & $54_{20.0}$ & $60_{20.0}$ & $63_{20.0}$ & $56_{20.0}$ \\
        \rowcolor{gray!20}
        \cellcolor{white}
        & CAA & $67_{2.0, 4}$ & $64_{2.0, 4}$ & $61_{2.0, 0}$ & $56_{2.0, 6}$ & $63_{1.5, 12}$ & $56_{2.0, 6}$ & $69_{2.0, 4}$ & $63_{2.0, 0}$ & $60_{2.0, 0}$ \\

        \multirow[t]{3}{*}{Pythia-1b} & No int. & $44$ & $44$ & $44$ & $44$ & $44$ & $44$ & $44$ & $44$ & $44$ \\
        & ITI & $54_{20.0}$ & $54_{20.0}$ & $54_{20.0}$ & $54_{20.0}$ & $54_{20.0}$ & $54_{20.0}$ & $54_{20.0}$ & $54_{20.0}$ & $54_{20.0}$ \\
        \rowcolor{gray!20}
        \cellcolor{white}
        & CAA & $59_{2.0, 8}$ & $62_{2.0, 5}$ & $54_{2.0, 0}$ & $57_{2.0, 4}$ & $59_{2.0, 10}$ & $56_{2.0, 4}$ & $57_{2.0, 3}$ & $60_{2.0, 5}$ & $54_{2.0, 0}$ \\

        \multirow[t]{3}{*}{Pythia-6.9b} & No int. & $44$ & $44$ & $44$ & $44$ & $44$ & $44$ & $44$ & $44$ & $44$ \\
        & ITI & $45_{20.0}$ & $54_{20.0}$ & $44_{0.0}$ & $54_{20.0}$ & $54_{20.0}$ & $54_{20.0}$ & $54_{20.0}$ & $54_{20.0}$ & $54_{20.0}$ \\
        \rowcolor{gray!20}
        \cellcolor{white}
        & CAA & $56_{1.5, 12}$ & $71_{1.5, 9}$ & $55_{2.0, 23}$ & $55_{2.0, 4}$ & $63_{1.5, 11}$ & $55_{2.0, 4}$ & $55_{2.0, 23}$ & $71_{1.5, 9}$ & $55_{2.0, 23}$ \\

        \multirow[t]{3}{*}{Pythia-6.9b-chat} & No int. & $55$ & $54$ & $28$ & $36$ & $64$ & $20$ & $44$ & $67$ & $30$ \\
        & ITI & $57_{15.0}$ & $54_{0.0}$ & $28_{0.0}$ & $44_{15.0}$ & $71_{15.0}$ & $32_{15.0}$ & $44_{0.0}$ & $67_{0.0}$ & $30_{0.0}$ \\
        \rowcolor{gray!20}
        \cellcolor{white}
        & CAA & $68_{1.5, 15}$ & $65_{1.5, 12}$ & $57_{1.5, 11}$ & $54_{1.5, 10}$ & $75_{1.5, 5}$ & $48_{1.5, 10}$ & $58_{1.5, 15}$ & $67_{0.0, 0}$ & $54_{1.5, 10}$ \\

        \multirow[t]{3}{*}{Pythia-12b} & No int. & $44$ & $44$ & $44$ & $44$ & $44$ & $44$ & $44$ & $44$ & $44$ \\
        & ITI & $54_{20.0}$ & $54_{20.0}$ & $54_{20.0}$ & $54_{20.0}$ & $54_{20.0}$ & $54_{20.0}$ & $54_{20.0}$ & $54_{20.0}$ & $54_{20.0}$ \\
        \rowcolor{gray!20}
        \cellcolor{white}
        & CAA & $54_{2.0, 0}$ & $64_{2.0, 9}$ & $54_{2.0, 0}$ & $60_{2.0, 11}$ & $58_{2.0, 11}$ & $55_{2.0, 12}$ & $54_{2.0, 0}$ & $67_{2.0, 10}$ & $54_{2.0, 0}$ \\

        \bottomrule
    \end{tabular}
    }
    \caption{Activation intervention: comparison between ITI \citep{li2023inferencetime} and CAA \citep{rimsky2023steering}. For ITI, the subscript indicates the value of the coefficient $\alpha_{\mathrm{ITI}}$ used: $\mathrm{Acc}_{\alpha_{\mathrm{ITI}}}$. For CAA, the subscript indicates first the value of the coefficient $\alpha$ used and second the layer $l$ at which intervention takes place: $\mathrm{Acc}_{\alpha_{\mathrm{CAA}}, l}$.}
    \label{tab:caa_params}
\end{table*}

\begin{table*}[t]
    \centering
    \small
    \begin{tabular}{lllc} 
        \toprule
        \textbf{Model} & \textbf{Method} & \textbf{Control} & \textbf{CAA Parameters} \\
        \midrule

        \multirow[t]{3}{*}{Llama-2-7b} & No int. & $44$ & \\
        \rowcolor{gray!20}
        \cellcolor{white}
        & CAA & $66_{\textcolor{PineGreen}{+22}}$ & $2.0$, $11$ \\

        \multirow[t]{3}{*}{Llama-2-7b-chat} & No int. & $56$ & \\
        \rowcolor{gray!20}
        \cellcolor{white}
        & CAA & $70_{\textcolor{PineGreen}{+14}}$ & $1.0$, $11$ \\

        \multirow[t]{3}{*}{Llama-2-13b} & No int. & $52$ & \\
        \rowcolor{gray!20}
        \cellcolor{white}
        & CAA & $85_{\textcolor{PineGreen}{+33}}$ & $2.0$, $12$ \\

        \multirow[t]{3}{*}{Llama-2-13b-chat} & No int. & $84$ & \\
        \rowcolor{gray!20}
        \cellcolor{white}
        & CAA & $97_{\textcolor{PineGreen}{+13}}$ & $1.0$, $12$ \\

        \multirow[t]{3}{*}{Llama-2-70b} & No int. & $90$ & \\
        \rowcolor{gray!20}
        \cellcolor{white}
        & CAA & $99_{\textcolor{PineGreen}{+9}}$ & $2.0$, $16$ \\

        \multirow[t]{3}{*}{Llama-2-70b-chat} & No int. & $69$ & \\
        \rowcolor{gray!20}
        \cellcolor{white}
        & CAA & $92_{\textcolor{PineGreen}{+23}}$ & $1.5$, $18$ \\

        \multirow[t]{3}{*}{Pythia-70m} & No int. & $41$ & \\
        \rowcolor{gray!20}
        \cellcolor{white}
        & CAA & $62_{\textcolor{PineGreen}{+21}}$ & $1.0$, $2$ \\

        \multirow[t]{3}{*}{Pythia-410m} & No int. & $48$ & \\
        \rowcolor{gray!20}
        \cellcolor{white}
        & CAA & $67_{\textcolor{PineGreen}{+19}}$ & $2.0$, $4$ \\

        \multirow[t]{3}{*}{Pythia-1b} & No int. & $44$ & \\
        \rowcolor{gray!20}
        \cellcolor{white}
        & CAA & $59_{\textcolor{PineGreen}{+15}}$ & $2.0$, $8$ \\

        \multirow[t]{3}{*}{Pythia-6.9b} & No int. & $44$ & \\
        \rowcolor{gray!20}
        \cellcolor{white}
        & CAA & $56_{\textcolor{PineGreen}{+12}}$ & $1.5$, $12$ \\

        \multirow[t]{3}{*}{Pythia-6.9b-chat} & No int. & $55$ & \\
        \rowcolor{gray!20}
        \cellcolor{white}
        & CAA & $68_{\textcolor{PineGreen}{+13}}$ & $1.5$, $15$ \\

        \multirow[t]{3}{*}{Pythia-12b} & No int. & $44$ & \\
        \rowcolor{gray!20}
        \cellcolor{white}
        & CAA & $54_{\textcolor{PineGreen}{+10}}$ & $2.0$, $0$ \\

        \bottomrule
    \end{tabular}
    \caption{Results for CAA \cite{rimsky2023steering} on the Forward Belief True Control task in BigToM \cite{gandhi2024understanding}. Numbers indicate accuracy scores, with differences (CAA $-$ No int.) as subscripts.}
    \label{tab:caa_control}
\end{table*}

\subsection{Compute resources}
We ran our experiments on a server running Ubuntu 22.04, equipped with eight NVIDIA Tesla V100-SXM2 GPUs with 32GB of memory and Intel Xeon Platinum 8260 CPUs. 

\subsection{Code}
Our code is provided as supplementary material and it will be made public under the MIT licence at \url{www.this-is-a-placeholder.com}.

\subsection{Societal impact}
While our work is foundational and remains distant from specific applications with direct societal impact, it's important to recognise the ethical implications of predicting and editing mental state representations.

Handling sensitive aspects of individuals' inner experiences and emotions requires careful consideration to avoid reinforcing biases or misunderstanding psychological nuances.
As LMs begin to encode aspects of ToM, there's a risk that over-interpreting these capabilities could lead to misplaced trust -- especially in real-world applications requiring nuanced social reasoning, such as education, healthcare, or mental health support.

Furthermore, while techniques like CAA show promise for steering internal representations, they also potentially introduce new ethical challenges. 
Manipulating a model's internal states, especially in ways that affect social reasoning, requires transparency and caution to avoid unintended consequences such as bias amplification or fairness issues.
Future work should consider not only improving technical performance but also developing safeguards and evaluation frameworks to ensure responsible use of ToM-like abilities in LMs.

\onecolumn
\onecolumn
\begin{example}{box:fb}{CadetBlue}{Forward Belief}
    \texttt{Noor is working as a barista at a busy coffee shop. Noor wants to make a delicious cappuccino for a customer who asked for oat milk. Noor grabs a milk pitcher and fills it with oat milk. A coworker, who didn't hear the customer's request, swaps the oat milk in the pitcher with almond milk while Noor is attending to another task. \textcolor{ablue}{Noor sees her coworker swapping the milk.} \textcolor{ared}{Noor does not see her coworker swapping the milk.} \\
    Does Noor believe the milk pitcher contains oat milk or almond milk? \\
    a) Noor believes the milk pitcher contains oat milk. \\
    b) Noor believes the milk pitcher contains almond milk.    
    }
\end{example}

\begin{example}{box:fa}{CadetBlue}{Forward Action}
    \texttt{Noor is working as a barista at a busy coffee shop. Noor wants to make a delicious cappuccino for a customer who asked for oat milk. Noor grabs a milk pitcher and fills it with oat milk. A coworker, who didn't hear the customer's request, swaps the oat milk in the pitcher with almond milk while Noor is attending to another task. \textcolor{ablue}{Noor sees her coworker swapping the milk.} \textcolor{ared}{Noor does not see her coworker swapping the milk.} \\
    What will Noor do? \\
    a) Noor will make the cappuccino using the milk in the pitcher. \\
    b) Noor will open the fridge once again to take out the oat milk and replace the almond milk with oat milk.}
\end{example}

\begin{example}{box:bb}{CadetBlue}{Backward Belief}
    \texttt{Noor is working as a barista at a busy coffee shop. Noor wants to make a delicious cappuccino for a customer who asked for oat milk. Noor grabs a milk pitcher and fills it with oat milk. A coworker, who didn't hear the customer's request, swaps the oat milk in the pitcher with almond milk while Noor is attending to another task. \textcolor{ablue}{Noor opens the fridge again and reaches for the oat milk.} \textcolor{ared}{Noor makes the cappuccino using the milk in the pitcher.} \\
    Does Noor believe the milk pitcher contains oat milk or almond milk? \\
    a) Noor believes the milk pitcher contains oat milk. \\
    b) Noor believes the milk pitcher contains almond milk.
    }
\end{example}

\begin{example}{box:default}{darkgray}{Default prompt}
    \texttt{Story: Noor is working as a barista at a busy coffee shop. Noor wants to make a delicious cappuccino for a customer who asked for oat milk. Noor grabs a milk pitcher and fills it with oat milk. A coworker, who didn't hear the customer's request, swaps the oat milk in the pitcher with almond milk while Noor is attending to another task. Noor does not see her coworker swapping the milk.\\
    Belief: Noor believes the milk pitcher contains oat milk.}
\end{example}

\begin{example}{box:random}{darkgray}{Prompt variation -- Random}
    \texttt{Story: Noor is working as a barista at a busy coffee shop. Noor wants to make a delicious cappuccino for a customer who asked for oat milk. Noor grabs a milk pitcher and fills it with oat milk. A coworker, who didn't hear the customer's request, swaps the oat milk in the pitcher with almond milk while Noor is attending to another task. Noor does not see her coworker swapping the milk.\\
    Belief: \textcolor{Bittersweet}{장士 decided [\_ countries sections behoSOUR gminy bef} Noor believes the milk pitcher contains oat milk.}
\end{example}

\begin{example}{box:misleading}{darkgray}{Prompt variation -- Misleading}
    \texttt{Story: Noor is working as a barista at a busy coffee shop. Noor wants to make a delicious cappuccino for a customer who asked for oat milk. Noor grabs a milk pitcher and fills it with oat milk. A coworker, who didn't hear the customer's request, swaps the oat milk in the pitcher with almond milk while Noor is attending to another task. Noor does not see her coworker swapping the milk.\\
    Belief: Noor believes the milk pitcher contains oat milk. \\
    \textcolor{Bittersweet}{Belief: The water valve is closed.}}
\end{example}

\begin{example}{box:time}{darkgray}{Prompt variation -- Time specification}
    \texttt{Story: Noor is working as a barista at a busy coffee shop. Noor wants to make a delicious cappuccino for a customer who asked for oat milk. Noor grabs a milk pitcher and fills it with oat milk. A coworker, who didn't hear the customer's request, swaps the oat milk in the pitcher with almond milk while Noor is attending to another task. Noor does not see her coworker swapping the milk.\\
    Belief: \textcolor{Bittersweet}{In the end,} Noor believes the milk pitcher contains oat milk.}
\end{example}

\begin{example}{box:init}{darkgray}{Prompt variation -- Initial belief}
    \texttt{Story: Noor is working as a barista at a busy coffee shop. Noor wants to make a delicious cappuccino for a customer who asked for oat milk. Noor grabs a milk pitcher and fills it with oat milk. \textcolor{Bittersweet}{Noor believes that the milk pitcher contains oatmilk.} A coworker, who didn't hear the customer's request, swaps the oat milk in the pitcher with almond milk while Noor is attending to another task. Noor does not see her coworker swapping the milk.\\
    Belief: Noor believes the milk pitcher contains oat milk.}
\end{example}

\begin{example}{box:eval}{teal}{Prompt used for downstream evaluation}
    \texttt{Answer the questions based on the context. Keep your answer concise, few words are enough, maximum one sentence. Answer as 'Answer:<option>)<answer>'.\\
    \\
    Story: Noor is working as a barista at a busy coffee shop. Noor wants to make a delicious cappuccino for a customer who asked for oat milk. Noor grabs a milk pitcher and fills it with oat milk. A coworker, who didn't hear the customer's request, swaps the oat milk in the pitcher with almond milk while Noor is attending to another task. Noor does not see her coworker swapping the milk.\\
    Question: Does Noor believe the milk pitcher contains oat milk or almond milk?\\
    Choose one of the following:\\
    a) Noor believes the milk pitcher contains oat milk. \\
    b) Noor believes the milk pitcher contains almond milk. \\
    Answer:}
\end{example}

\twocolumn

\end{document}